%% file: main.tex
\documentclass[manuscript,screen,nonacm]{acmart} 
\AtBeginDocument{%
  }

\setcopyright{cc}
\setcctype[4.0]{by}
\copyrightyear{2026}

\acmConference[Conference acronym 'XX]{Make sure to enter the correct
  conference title from your rights confirmation email}{June 03--05,
  2018}{Woodstock, NY}
\acmISBN{978-1-4503-XXXX-X/2018/06}

\usepackage{colortbl}
\usepackage{array}

\usepackage{enumitem}
\usepackage[normalem]{ulem}

\usepackage[table]{xcolor}
\usepackage[export]{adjustbox}

\usepackage{amssymb}

\usepackage{tabularx}
\usepackage{booktabs}
\usepackage{array}
\usepackage[export]{adjustbox}

\definecolor{significant}{HTML}{93C47D}
\definecolor{nonsignificant}{HTML}{E0E0E0}

\begin{document}

\title[Socially Assistive Robot-Delivered Cognitive Behavioral Therapy Exercises]{LLM-Powered Socially Assistive Robot-Delivered Cognitive Behavioral Therapy Exercises: an Exploratory Study with University Students}

\author{Mina Kian}
\authornote{Corresponding author.}
\email{kian@usc.edu}
\affiliation{%
  \institution{Interaction Lab, Viterbi School of Engineering, University of Southern California}
  \city{Los Angeles}
  \state{California}
  \country{USA}
}

\author{Mingyu Zong}
\authornote{These authors contributed equally to this work.}
\affiliation{%
  \institution{Autonomous Networks Research Group, Viterbi School of Engineering, University of Southern California}
  \city{Los Angeles}
  \state{California}
  \country{USA}
}

\author{Katrin Fischer}
\authornotemark[2]
\affiliation{%
  \institution{Interaction Lab, Viterbi School of Engineering, University of Southern California}
  \city{Los Angeles}
  \state{California}
  \country{USA}
}

\author{Abhyuday Singh}
\authornotemark[2]
\affiliation{%
  \institution{Interaction Lab, Viterbi School of Engineering, University of Southern California}
  \city{Los Angeles}
  \state{California}
  \country{USA}
}

\author{Anna-Maria Velentza}
\authornotemark[2]
\affiliation{%
  \institution{Brest National School of Engineering}
  \city{Plouzané}
  \country{France}
}

\author{Amy O'Connell}
\affiliation{%
  \institution{Interaction Lab, Viterbi School of Engineering, University of Southern California}
  \city{Los Angeles}
  \state{California}
  \country{USA}
}

\author{Lydia Ignatova}
\affiliation{%
  \institution{Interaction Lab, Viterbi School of Engineering, University of Southern California}
  \city{Los Angeles}
  \state{California}
  \country{USA}
}

\author{Pau Sang}
\affiliation{%
  \institution{Interaction Lab, Viterbi School of Engineering, University of Southern California}
  \city{Los Angeles}
  \state{California}
  \country{USA}
}

\author{Shriya Upadhyay}
\affiliation{%
  \institution{Interaction Lab, Viterbi School of Engineering, University of Southern California}
  \city{Los Angeles}
  \state{California}
  \country{USA}
}

\author{Anika Gupta}
\affiliation{%
  \institution{Interaction Lab, Viterbi School of Engineering, University of Southern California}
  \city{Los Angeles}
  \state{California}
  \country{USA}
}

\author{Misha A. Faruki}
\affiliation{%
  \institution{Interaction Lab, Viterbi School of Engineering, University of Southern California}
  \city{Los Angeles}
  \state{California}
  \country{USA}
}

\author{Wallace Browning}
\affiliation{%
  \institution{Interaction Lab, Viterbi School of Engineering, University of Southern California}
  \city{Los Angeles}
  \state{California}
  \country{USA}
}

\author{S\'ebastien M. R. Arnold}
\affiliation{%
  \institution{Interaction Lab, Viterbi School of Engineering, University of Southern California}
  \city{Los Angeles}
  \state{California}
  \country{USA}
}

\author{Meng Chen}
\affiliation{%
  \institution{Department of Psychology, University of Southern California}
  \city{Los Angeles}
  \state{California}
  \country{USA}
}

\author{Bhaskar Krishnamachari}
\affiliation{%
  \institution{Autonomous Networks Research Group,  Viterbi School of Engineering, University of Southern California}
  \city{Los Angeles}
  \state{California}
  \country{USA}
}

\author{Maja J. Matari\'c}
\affiliation{%
  \institution{Interaction Lab, Viterbi School of Engineering, University of Southern California}
  \city{Los Angeles}
  \state{California}
  \country{USA}
}

\renewcommand{\shortauthors}{Kian et al.}

\begin{abstract}
Mental health is a significant healthcare challenge, and cognitive behavioral therapy (CBT) is a widely used therapeutic method for treating anxiety and depression. However, traditional CBT often requires access to trained clinicians and can be cost-prohibitive or logistically difficult for many individuals. To address these barriers, we developed a low-cost socially assistive robot (SAR) that uses a large language model (LLM) to guide the user through interactive at-home CBT exercises. In this exploratory study, 38 university students completed CBT exercises across a 15-day period using one of three modalities: with a robot (using an LLM for dialogue), a chatbot (using the same LLM for dialogue), or traditional CBT worksheets. We measured weekly therapeutic outcomes, changes in pre-/post-session anxiety measures, and adherence to completing CBT exercises. Our findings indicate that self-reported general psychological distress significantly decreased over the study period in the robot and worksheet conditions but not in the chatbot condition. Additionally, the SAR enabled significant single-session improvements on more days than the other two conditions combined. Mixed-effects modeling further suggested that the robot and chatbot conditions better reduced post-session anxiety for those with elevated levels of anxiety. Our findings suggest that SAR-guided, LLM-powered CBT may be an effective method for supporting therapeutic progress and decreasing user anxiety immediately after completing the CBT exercise. The findings underscore the potential for combining AI-driven personalization with socially assistive robotics to create accessible, scalable, and engaging mental health interventions.
\end{abstract}



\keywords{human robot interaction (HRI), socially assistive robotics (SAR), cognitive behavioral therapy (CBT), mental health, large-language model (LLM), longitudinal study, university students}

\received{03 February 2026}
\received[revised]{16 June 2026}
\received[accepted]{22 June 2026}
\received[published]{09 September 2026}

\maketitle
\begin{center}
\footnotesize
© 2026 Kian, Zong, Fischer, Singh, Velentza, O’Connell, Ignatova,
Sang, Upadhyay, Gupta, Faruki, Browning, Arnold, Chen,
Krishnamachari, and Matarić.

This work is licensed under the
\href{https://creativecommons.org/licenses/by/4.0/}
{Creative Commons Attribution 4.0 International License}.
This is an author-formatted version of the article published in
\textit{Frontiers in Robotics and AI}.
The version of record is available at
\href{https://doi.org/10.3389/frobt.2026.1802622}
{https://doi.org/10.3389/frobt.2026.1802622}.
\end{center}

\input{sections/intro.tex}

\input{sections/methods.tex}

\input{sections/results.tex}

\input{sections/discussion.tex}

\input{sections/statements.tex}


\bibliographystyle{ACM-Reference-Format}
\bibliography{references}

\appendix
\input{sections/appendix.tex}

\end{document}

%% file: sections/intro.tex
\section{Introduction}\label{sec1}

University students face increased periods of vulnerability in mental health and consequently experience anxiety, depression, and suicidality~\citep{liu2019prevalence}. 
At the same time, mental health services are often inaccessible to students on university campuses due to high demand. 
A survey conducted across 108 U.S. institutions indicated an urgent need for mental health support, with stress and suicidality especially pronounced among racial, ethnic, sexual, and gender minorities~\citep{liu2019prevalence}. 

Cognitive Behavioral Therapy (CBT) is an effective approach for addressing mental health challenges~\citep{beck2020cognitive} and is a widely practiced, clinically validated form of therapy~\citep{hofmann2012efficacy}.
The premise of CBT is that ``psychological problems are based, in part, on faulty or unhelpful ways of thinking [...] and learned patterns of unhelpful behavior'' and that people can learn better ways to cope, enabling them to be more effective in their lives~\citep{whatiscognitivebehavioraltherapy_2017}.
CBT treatment guides people to recognize and address their distorted thinking and alter their thought patterns~\citep{kuru2018cognitive} as well as learn to adjust their behavior to influence their emotional state ~\citep{boswell2017behavioral}.

For CBT to be effective, those receiving it must practice applying CBT strategies in their daily lives. CBT therapists often assign \textit{homework}~\citep{beck1993cbt, dobson2021commentary}, exercises to be completed at home that create opportunities for regular practice ~\citep{prasko2022homework} and encourage independence in addressing unhelpful patterns of thought and behavior~\citep{kazantzis2021introduction}.  Such homework typically involves completing written worksheets that guide the user through a set of steps that apply a particular CBT strategy to a recent experience~\citep{beck1993cbt}. 
Worksheets help users work through thoughts via clear written steps, but they lack the interactivity of conversations with a therapist 
and instead provide static instructions; their effectiveness therefore hinges on the user's willingness to share their thoughts. In contrast, in interactive discussions, a therapist's tailored follow-up questions can encourage greater reflection and new perspectives. 
Adherence to CBT homework is generally low, negatively affecting therapeutic outcomes~\citep{lebeau2013homework}.
Studies have shown that more frequent use of CBT techniques and practices in patients outside of therapy is associated with fewer symptoms of mental illnesses \citet{kazantzis2010meta}, including depression~\citep{hundt2013relationship}. 
In particular, ~\citet{kazantzis2010meta} established that CBT homework is a core mechanism of therapeutic change, with meta-analytic evidence demonstrating a reliable relationship between homework engagement and clinical outcomes.
Thus, investigating methods to support CBT homework completion is of interest to the field and motivates our work.

Innovations in natural language processing, in particular large language models (LLMs), have greatly expanded the capabilities of dialogue-based therapeutic technologies. 
Applications of LLMs have grown rapidly, especially in conversational agents~\citep{zaib2020short}, and will grow further with improved task-specific controls~\citep{wei2022chain}.
For example, ~\citep{ziems2022inducing} found that LLMs can successfully complete CBT-based positive reframing tasks.
However, using LLMs for mental health has raised some concerns in the scientific and clinical communities because they may inadvertently generate potentially harmful or inappropriate messages ~\citep{lilianweng2023}; studies that employ LLMs must follow ethical and clinical guidelines to safeguard users from harm ~\citep{Schueller2023}.

While LLMs have greatly advanced human-machine dialogue, motivating and sustaining engagement in therapeutic exercises is a well-known challenge. Socially assistive robots (SARs), physically embodied agents designed to provide support through social interaction~\citep{feil2005defining}, have been shown to be effective in supporting motivation and practice for a wide range of health and therapeutic goals, including rehabilitation after stroke~\citep{swift2015effects, mataric2007socially}, 
 cerebral palsy~\citep{feingold2020social, malik2016emergence}), 
 cognitive and social skill training~\citep{scassellati2012robots, pennisi2016autism}, 
 encouraging physical activity~\citep{fasola2012using}, 
 mental health outcomes ~\citep{kabacinska2021socially, scoglio2019use, guemghar2022social}, 
 and other areas of support and behavior change~\citep{scassellati2012robots, ferrante2021social, robinson2020social}. 
 In a systematic review, \citet{deng2019embodiment} concluded that physically embodied agents, i.e., robots, demonstrate performance improvements over virtual agents in a broad range of settings, particularly for relationship-oriented tasks where social presence and engagement are critical for maintaining rapport ~\citep{deng2019embodiment, segura2012you}.
A longitudinal HRI study found that, even as the novelty of interacting with a robot wore off, users reported increased interaction quality and perceived communication competence ~\citep{guy2022longtermux}.  
Studies indicate that children and adults alike are willing to share confidential or embarrassing information with a robot, often to the same extent they would with each other~\citep{bethel2011secret, barendregt2014child, pitardi2021service}.
This greater social presence allows robots to offer better encouragement and guidance to users in challenging tasks~\citep{mataric2016socially}.

Combining the strengths of SAR and LLMs is therefore a natural direction for exploration. LLMs have enabled SARs to engage in more natural, context-aware dialogue with users across domains~\citep{brown2020language, zhao2023survey}. Several recent SAR studies incorporate LLMs to support more personalized interactions~\citep{shi2024can, abdel2024enhancing, lima2025promoting}. Our study explores how an LLM-enabled SAR may support users in completing interactive CBT homework exercises in their daily lives.
This work seeks to leverage the benefits of physical embodiment by developing an LLM-powered in-home SAR to encourage individuals to complete daily CBT homework exercises. 

Prior to bringing SARs and LLMs together, studies have explored each technology separately in the CBT context. 
For example, \citet{dino2019delivering} showed that elderly users receiving CBT from a SAR enjoyed interacting with the robot, had elevated moods, and found the CBT information valuable, but this experiment was limited by its use of a fixed pre-programmed dialogue tree.
Stronger conversational agents, including over 40 chatbots, have been developed for therapeutic purposes such as administering therapy, training therapists, screening for mental health symptoms, and encouraging self-management~\citep{abd2019overview}. Chatbots have been shown to reduce some symptoms of depression and anxiety ~\citep{fulmer2018using, ahmed2023chatbot}.
Recent work has developed chatbots with greater personalization and interactivity, aiming to enhance accessibility, adherence, and therapeutic outcomes ~\citep{fitzpatrick2017delivering, burton2016pilot, denecke2020mental}.
For example, \citet{kumar2022exploring} developed a constrained language model-powered chatbot that guided participants through 5-minute therapy sessions,  modulating the chatbot’s identity, behavior, and personality to test participant preferences. 
Similarly, \citet{jang2021mobile} showed that a chatbot delivering CBT and psychoeducation for adults with ADHD could be effective in relieving their symptoms.
Technology-based interventions have been shown to be beneficial because of their fast response times compared to human therapists, and their ability to bridge the vocabulary gap between therapists and patients~\citep{kumar2016sanative}.

In this work, we explored the combined promise of LLMs and SARs to assess their effectiveness in CBT homework exercise delivery.
We conducted a 15-day exploratory user study with a sample of university students (N=38) to understand how they respond to LLM-powered and SAR-guided CBT homework exercises.
In this mixed design between- and within-subjects study, participants were assigned to one of three conditions and accordingly completed at-home CBT homework exercises using a SAR, chatbot, or traditional worksheets, respectively. 
We collected qualitative feedback from participants about their experience and measured the daily and weekly therapeutic outcomes in each condition.
In the last 7 days of the study, completing the homework exercises was optional, allowing us to observe patterns of user adherence to each CBT delivery method.
We found that, over the study period, psychological distress significantly decreased in the robot and worksheet conditions.
We also found that pre- to post-session anxiety significantly decreased on more days in the robot condition than in the chatbot and worksheet conditions.
Our findings provide novel insights about the effectiveness of LLM-powered SAR-guided CBT and its impact on therapeutic outcomes.

\noindent This work aimed to address the following research questions with university students as the intended target (not convenience) user population:

\begin{enumerate}[
    label=\textbf{RQ\arabic*.},
    leftmargin=*
]
    \item How does the CBT in-home exercise delivery method affect immediate and short-term therapeutic outcomes?

    \item How does the CBT in-home exercise delivery method affect adherence to regular CBT practice?

    \item How does the CBT in-home exercise delivery method affect students' perception and experience with the system?

    \begin{enumerate}[
        label=\textbf{RQ3\alph*.},
        leftmargin=*
    ]
        \item How do students respond to the platform structure as a whole?

        \item How do students respond to LLM-driven CBT homework exercises?

        \item How do students respond to robot-delivered CBT homework exercises?
    \end{enumerate}
\end{enumerate}

%% file: sections/methods.tex
\section{Methodology}\label{sec1}

This study examined the effectiveness of CBT homework delivered through three modalities: (1) traditional worksheets completed without an agent; (2) a text-based chatbot powered by an LLM; and (3) a socially assistive robot (SAR) powered by the same LLM. We collected quantitative measures of therapeutic outcomes, anxiety, and adherence, as well as qualitative feedback from post-study interviews, to provide a comprehensive view of participants’ experiences. The following section describes the system design, study procedures, participant recruitment, data collection instruments, and analysis methods.

\subsection{System Design}

Standard CBT exercises were adapted for delivery through three distinct modalities: traditional worksheets, a text-based chatbot, and a SAR. We used GPT-3.5, OpenAI’s most advanced model at the time, to deliver CBT exercises in the chatbot and SAR modalities. Earlier pilot testing was conducted with GPT-3, as GPT-3.5 had not yet been released.

\subsubsection{CBT Exercise Development}\label{prompt_generation}
CBT homework worksheets offer a structured approach to CBT exercises, with a primary goal of improving patients' skill acquisition.
We examined several online databases of CBT homework worksheets that were designed by CBT therapists using CBT principles and intended for use by CBT therapists~\footnote{https://www.therapistaid.com/; https://thinkcbt.com/think-cbt-worksheets; https://positivepsychology.com/cbt-cognitive-behavioral-therapy-techniques-worksheets/\#6-evidence-based-cbt-worksheets}.
From these worksheets, we selected a subset of exercises that readily translated into interactive exercises guided by an agent (robot or chatbot, depending on the study condition). CBT worksheets for the same exercise vary in structure and phrasing, and when delivered by a clinician in a session, they are often dynamically adapted to meet the client’s needs. Our LLM-based approach aimed to achieve the goals of active CBT practice and homework completion, since more frequent CBT practice has been linked with fewer symptoms of mental illness \citep{kazantzis2010meta}.

\paragraph{Prompt Generation for LLM Delivery}
To adapt the exercises for LLM delivery, we developed a series of prompts that allowed an LLM to guide users through each exercise. These prompts were refined and informally evaluated through in-lab pilot testing. In addition to prompt engineering, we compared the LLM's performance with zero-shot and few-shot learning examples.
In the zero-shot learning setup, the LLM-powered agent received a description of the exercise and its role in the conversation to support the exercise, with no example conversations provided. 
In the few-shot learning setting, we additionally provided the LLM with 2-3 example conversations demonstrating how the CBT exercise should be conducted.
After testing, we found that zero-shot learning with an appropriate prompt was enough for the model to properly and effectively deliver the exercises.
With limited input tokens, it was challenging to select representative few-shot examples that enabled the agent to respond appropriately when the conversation topic differed from the example dialogue. We found that the model frequently repeated the language from the examples, regardless of the present context.
We therefore adopted a zero-shot learning approach for the study.
Additionally, omitting examples allowed the model to make use of the full session conversation history without reaching the token limit, enabling it to better adhere to the context.
Research group members tested a variety of prompts corresponding to the list of candidate CBT exercises and identified the three best-performing exercises based on how closely the resulting interactions resembled a therapist-led session in terms of structure, flow, and clinical appropriateness. 

\paragraph{Prompt Validation}

Through an in-lab pilot study with 15 participants, we established that the LLM was able to effectively and safely manage two of the tested CBT exercises: Cognitive Restructuring~\footnote{Worksheet source: Cognitive Restructuring: Socratic Questions, https://www.therapistaid.com/therapy-worksheet/socratic-questioning/cbt/none\label{fn:cognitive_restructuring_worksheet}}~\citep{clark2015socratic} and Coping Strategies~\footnote{Worksheet source: Coping Styles Formulation, https://positive.b-cdn.net/wp-content/uploads/Coping-Styles-Formulation.pdf\label{fn:coping_strategies_worksheet}}~\citep{kuyken2008science}.
Cognitive Restructuring supports identifying and altering cognitive distortions and negative/maladaptive thinking patterns~\citep{clark2015socratic}.
Coping Strategies is based on the CBT case conceptualization framework, helping individuals better understand the problems they face and identify healthy and productive ways to manage them~\citep{kuyken2008science}. 

The final prompts used in both the chatbot and SAR conditions are provided in Table~\ref{tab:prompts} (model parameters are listed in Table~\ref{tab:model_parameters} in the Supplementary Materials).

\begin{table}[htbp]
\centering
\caption{\textbf{LLM Prompts.} Model prompts used for the CBT exercises and the moderator.}
\label{tab:prompts}

\begin{tabularx}{\linewidth}{@{}>{\bfseries}l X@{}}
\toprule
Prompt & Instructions \\
\midrule

Cognitive Restructuring &
Cognitive restructuring is a strategy to help the patient identify cognitive distortion and find evidence to challenge the distortion. You are a therapist who uses cognitive restructuring to help your patient in this session. \\

\addlinespace
Coping Strategies &
Coping strategy is used to help patients identify problems they encountered and the triggers. When a problem is defined, a therapist will help the patient figure out ways to cope with it. You are a therapist who uses coping strategies to help your patient in this session. \\

\addlinespace
Moderator &
``...Given the above context, is the following advice given by the therapist problematic? Reply with just a Yes or No.'' \\

\bottomrule
\end{tabularx}
\end{table}

\paragraph{Moderator Design}
Due to the sensitive nature of therapy, we used GPT-3.5 to design a moderator to check the LLM-generated responses and flag and filter out problematic statements. 
Although the chatbot did not produce any problematic responses during in-lab pilot testing, we used the moderator as an additional safety measure to minimize potential harm to study participants.
The GPT-3.5 moderator was designed using prompt engineering, as follows. We created over 200 examples of inappropriate therapeutic responses in various conversational contexts. 
We then tested various prompts (with both zero-shot and few-shot examples) to evaluate the resulting moderator's ability to recognize inappropriate responses. 
We selected the prompt with the best performance achieved through zero-shot learning. 
The selected moderator prompt is provided in Table \ref{tab:prompts} (for parameters, see Table \ref{tab:model_parameters} in Supplementary Materials).

\subsubsection{User Interface Design}

\begin{figure}[htbp]
\centering
\includegraphics[width=0.8\linewidth]{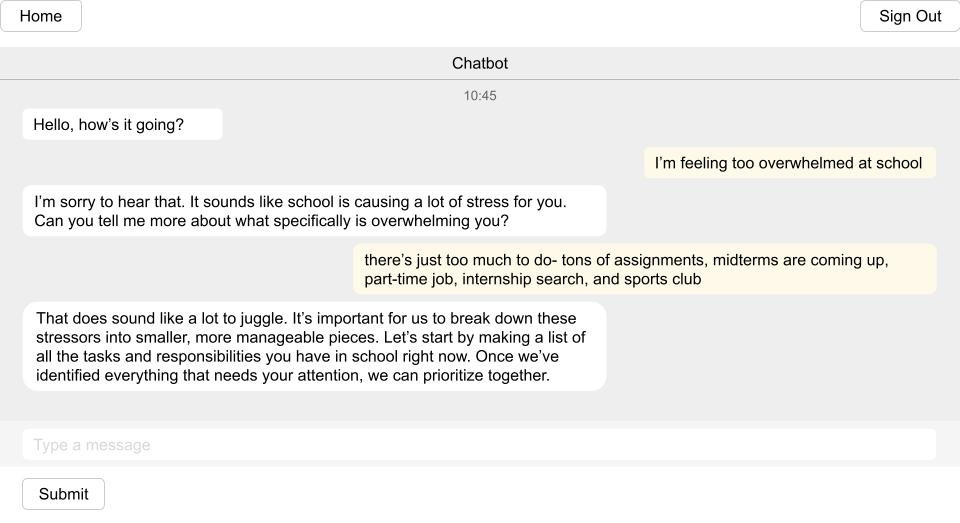}
\caption{\textbf{Web application interface.} The user interface for the CBT homework interaction. The interface was identical in the robot and chatbot conditions.}
\label{fig:chatbot_interface}
\end{figure}

\begin{figure}[htbp]
\centering
\includegraphics[width=0.6\linewidth]{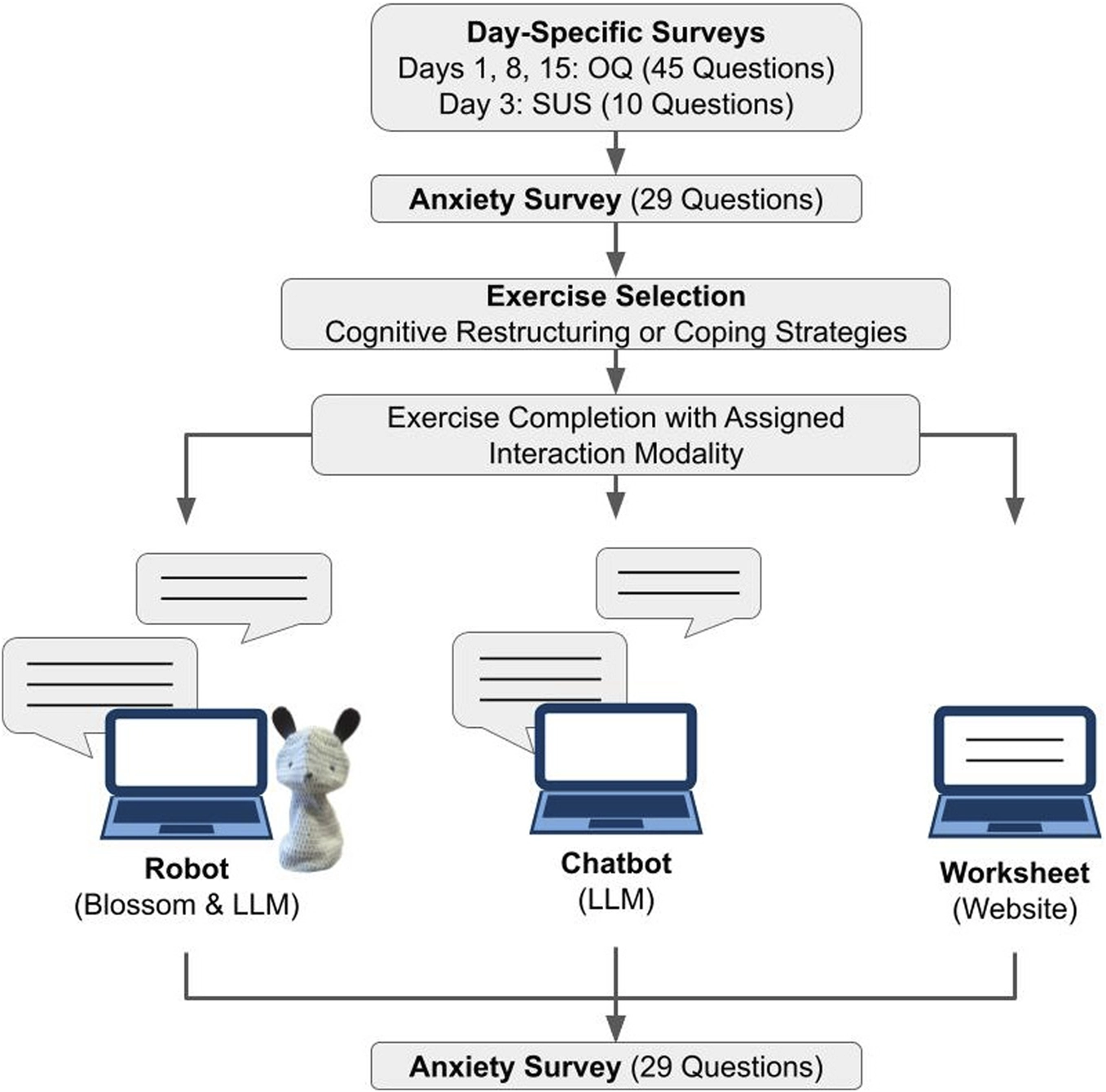}
\caption{\textbf{Interaction structure.} Flow chart showing the sequence of surveys and CBT exercise for all three conditions.}
\label{fig:cbt_session_structure}
\end{figure}

For the user interface, we designed a web application (Fig. \ref{fig:chatbot_interface}) that was securely hosted on Amazon Web Services (AWS), an IRB-approved cloud service where participants consented to have their study data stored. 
Access to the web application was password-protected; each participant was assigned unique credentials for authentication. 
At the start of each CBT homework session, participants completed an anxiety scale survey. 
Participants were then prompted to click a ``start'' button to begin recording video and audio, giving their consent to be recorded as they completed the CBT homework exercise.
They proceeded to complete the exercise through their assigned modality  (robot, chatbot, or worksheet, based on the study condition). 
In the worksheet condition, participants typed responses to the original questions from the selected CBT worksheets (Cognitive Restructuring\footref{fn:cognitive_restructuring_worksheet} or coping strategies\footref{fn:coping_strategies_worksheet}). 
All questions were displayed on a single, non-interactive web page.
In the robot and chatbot conditions, participants completed an interactive discussion with the LLM-powered agent by typing into a dialogue interface. 
The LLM-powered agent guided the participants through the exercise. 
In the robot condition, the robot verbalized the LLM response before it appeared on the screen.
(The LLM-powered moderator described above reviewed all LLM-based responses by the robot or chatbot to determine their appropriateness before they appeared on screen.) 
After finishing the CBT exercise, participants completed a post-session anxiety scale survey. 
At the end of the session, the website displayed the university's Student Health helpline phone number, in case participants needed to access further mental health support. 
The participants in the robot condition were given additional instructions on how to power down the robot. 
The data from each session were securely stored on IRB-approved cloud services.
The full session structure is shown in Fig. \ref{fig:cbt_session_structure}.

\renewcommand{\figurename}{Fig.}

\renewcommand{\figurename}{Fig.}

\subsubsection{Robot Design}
\begin{figure}[htbp]
\centering
\includegraphics[width=0.6\linewidth]{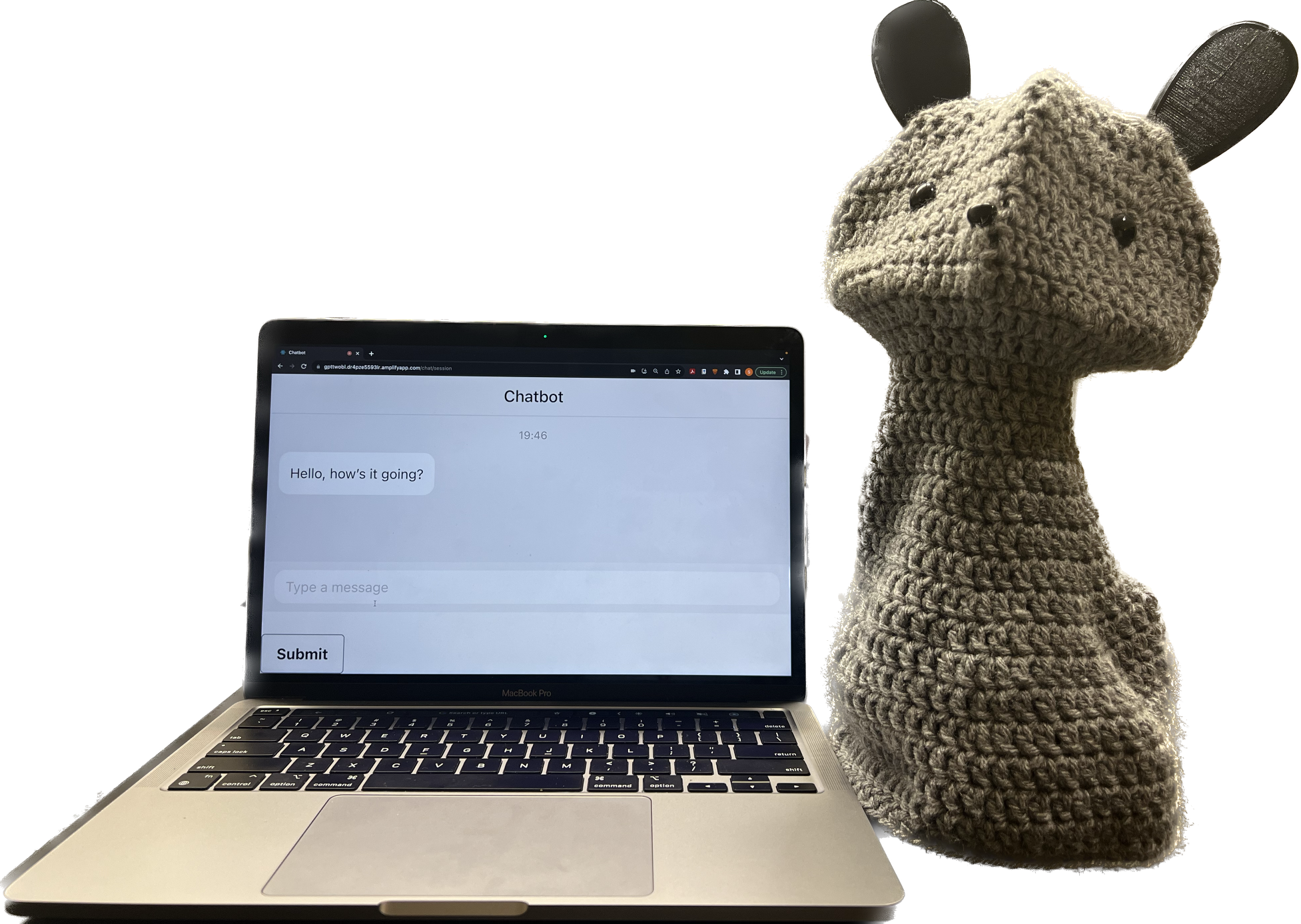}
\caption{\textbf{SAR design.} Blossom robot and laptop interface used in the robot condition in the study.}
\label{fig:Blossom_Robot_Design}
\end{figure}

To promote replicability and accessibility, we used the low-cost open-source Blossom socially assistive robot (see Figure \ref{fig:Blossom_Robot_Design})~\citep{suguitan2019blossom, shi2024build}. 
Each robot cost approximately 350 US dollars to build and was connected to a Raspberry Pi that cost approximately 100 US dollars.
The robot was covered with a neutral light gray crocheted exterior.
The robot used the AWS Polly female text-to-speech voice to verbalize responses. 
We chose a female voice because ~\citet{bhati2014effect} observed that both female and male clients reported higher levels of therapeutic alliance with female therapists than with male therapists. 

The robot had two states: observing and speaking. 
In the observing state, the robot's head continuously raised and lowered to mimic breathing, indicating that the robot was awake and attentive to the user.
We used the appearance of slow breathing to subtly encourage deep breathing relaxation by the user as it is often employed in therapeutic settings due to its efficacy in reducing stress~\citep{andrayani2023effectiveness}.
In the speaking state, the robot's head executed a gentle, oscillatory tilt from side to side, again indicating that it was awake and attentive~\citep{fink2012anthropomorphism}. 

In all three conditions, the participants provided their answers to the CBT exercise questions by typing them into the interface (or form, for the worksheet).  We chose this for several reasons.
We tested seven open-source speech-to-text (STT) packages and found that all had poor performance across various English accents representative of our university student population. 
In order to develop a system that was accessible to participants from a wide range of backgrounds, we chose to have participants enter their responses by typing instead of speaking.  This choice also added a measure of privacy to the participants since their responses to CBT homework questions could not be overheard.  Finally, it made all three experiment conditions more comparable by using the same response modality.

\subsection{Study Design}
We designed a 15-day between-subjects study with repeated measures to evaluate research outcomes.
Participants were split into three study conditions for completing the CBT homework exercises: 1) with an LLM-powered SAR, 2) with an LLM-powered chatbot, and 3) using traditional worksheets.
During the first half of the study (days 1-8), participants were required to complete daily CBT homework exercises through their assigned modality. 
The mandatory nature of the exercises intended to encourage the practice of completing the exercises daily as intended, and to collect data to compare outcomes among the three conditions.
In the second half of the study (days 9-15), participants were not required to complete daily CBT homework exercises.  The optional nature of the exercises intended to analyze how participants voluntarily interacted with their assigned delivery method. 
We used this approach because one objective of this study was to explore if the modality of delivering the CBT exercises impacts \textit{intentional} non-adherence~\citep{choi2015mobile, ingeson2018smart}.
To limit instances of unintentional non-adherence~\citep{abdulrahman2021search, kelders2012persuasive}, participants under all three conditions received daily emails throughout the study reminding them to complete the CBT homework exercise and questionnaires. 
Therapeutic outcomes were measured using the Anxiety Scale ~\citep{pilkonis2011item}, administered pre- and post-session as well as weekly measures administered on days 1, 8, and 15. The CBT exercise and surveys took approximately 30 minutes to complete. The usability of the interface was measured on Day 3. 

\subsection{Participants}

We recruited 42 student participants from a major US university to participate in our study, a cohort size primarily determined by the availability of 15 robots and the logistical challenges of coordinating multi-week at-home deployments.  This sample size is comparable to other in-home human-robot interaction SAR research studies, such as those by \citet{claybaugh2019asd} (N=17), \citet{jeong2020roboticpositivepsychologycoach} (N=35), \citet{matheus2022deepbreath} (N=43),   \citet{Jeong2022DeployingAR} (N=42), and \citet{OConnell2024DesignAE} (N=11), and aims to provide useful exploratory insights.

\subsubsection{Eligibility \& Ethical Screening}
This study was approved by the university's IRB. 
Participants were screened to be over 18 years of age, proficient in English, of normal or corrected-to-normal vision and hearing (due to the auditory and visual components of the study), and living near campus (to allow us to quickly address technical issues if any occurred).

After consulting the University's Institutional Review Board (IRB), we determined that, due to the sensitive nature of the CBT exercises, it was not appropriate to recruit severely depressed individuals in our sample of participants, and that recruiting severely depressed students to evaluate our previously untested system would pose an unnecessary risk to their safety and well-being. 
Participants were therefore screened using the PHQ-9 measure of depression~\citep{kroenke1999patient}; all responders who scored 15 or higher, indicating moderately severe to severe depression, were excluded from participating.

\subsubsection{Recruitment and Compensation}
Students were recruited through emails to major academic departments as well as an e-bulletin post on a study promotion board. All students who filled out the study interest form were given information on available campus mental health resources.
Participants were compensated with a US\$150.00 Amazon gift card after completing the study. 
The amount was calculated based on the amount of time we anticipated participants would spend on the study and the local minimum wage. Participants were informed that they would still be compensated in the full amount even if they chose not to complete any exercises in the optional week.

\subsubsection{Cohort Demographics}

After the study began, two participants in the chatbot condition and two participants in the worksheet condition discontinued their participation. Complete data were therefore collected from 38 participants: 14 in the robot condition, 12 in the chatbot condition, and 12 in the worksheet condition.

The 38 participants ranged from ages 18 to 27 (M = 20.57895, SD = 2.1), and self-reported their gender as 16 female, 18 male, 1 non-binary, and 3 preferred not to answer.
Ten of the participants were graduate students and 28 were undergraduates. 
Additional information about the participants is shown in {Fig. \ref{fig:race_distribution} of the Supplementary Materials}.

\subsection{Procedure}
Before the start of the data collection, each participant had an informed consent meeting with a researcher where they reviewed the study consent form and the requirements to maintain participation in the study. 
Participants were informed that for the first 8 days of the study, they would be required to complete at least one CBT homework exercise daily to continue participating. 
They were told that if they skipped the CBT homework exercises in the final 7 days of the study, it would not affect their participation in the study, and they would still receive full compensation. 
The researcher explained that the participants would receive a daily email reminding them to complete the CBT exercises and any required surveys every day of the study. 
The researcher set up the participants' login (and equipment for the robot condition) and guided them through a mock session to practice using the platform.
At the conclusion of the study, all participants completed a semi-structured post-study interview with a researcher lasting approximately one hour. During the interview, they reflected on the study and gave feedback on the platform they used.

\subsection{Data Collection and Instruments}
To address our research questions, we followed a triangular approach~\citep{ostlund2011combining}, collecting quantitative and qualitative data~\citep{babbie2020practice} to build a holistic understanding of participants' experiences with, and responses to, their respective CBT homework delivery method. 

The following survey instruments were used:

\indent{\it PHQ-9.} The Patient Health Questionnaire-9 is a 9-item self-report diagnostic tool that measures an adult patient's severity of depression on a 4-point Likert scale~\citep{kroenke1999patient}. Study candidates were excluded from participating if they scored 15 or higher. 

\indent{\it OQ.} The Outcomes Questionnaire-45 (OQ) is a 45-item instrument where individuals rate their experiences with a list of symptoms shared across multiple adult mental disorders and syndromes, on a 5-point Likert scale~\citep{lambert2004outcome}. The Full OQ score was administered on days 1, 8, and 15 to measure participants' therapeutic outcomes.

\indent{\it SUS.} The System Usability Scale (SUS) is a 10-item instrument that evaluates the ease of use of the system~\citep{brooke1996sus}. Each question is on a 5-point Likert scale. The full SUS was administered on Day 3, after participants had gained familiarity with the study platform.

\indent{\it Anxiety Scale.} Pilkonis et al.~\citep{pilkonis2011item} developed and calibrated an item bank for depression, anxiety, and anger as part of the Patient-Reported Outcomes Measurement Information System (PROMIS). The 29-item Calibrated Anxiety Items were administered immediately before and after each CBT homework session.

\indent {\it NARS.} The Negative Attitude towards Robots Scale is a 14-item instrument that measures users' attitudes and emotions towards robots~\citep{nomura2006experimental}. The NARS was administered on Day 1 of the study.

\indent{\it Semi-Structured Interview.}  We conducted a semi-structured interview with each participant after they completed the study to gather qualitative feedback about their experience using their assigned CBT homework delivery method. The interviews were conducted using a guide developed by the research team. Open-ended questions covered the following topics (platform refers to the robot/chatbot/worksheet): first impressions/beginning of the study, benefits of the platform for CBT homework, challenges with the platform, factors affecting their use of the platform for CBT homework, and overall experience during the study. The interview incorporated an introduction, a summary of the study's purpose, and a debriefing for participants to reflect on all aspects of the study and leave additional feedback. All interviews were conducted in English and recorded, with consent.

\subsection{Data Analysis}

\subsubsection{Statistical Analysis of Questionnaires}
Questionnaire data were analyzed using statistical tests in Python 3.11.3 libraries and R 4.3.1 packages, as described in Supplementary Materials. The dependent variables were the instrument scores (Anxiety Scale, Outcome Questionnaire, System Usability) and adherence. 
The instrument data were internally consistent, with Cronbach $\alpha$ $>$ 0.70. 
The independent variables were the study condition, NARS scores, and time.

\subsubsection{Therapeutic outcomes via weekly mental health assessments}
The Outcomes Questionnaire (OQ) was sent to the participants to measure their psychotherapy progress throughout the duration of the study.
Dependent t-tests were performed on the difference between the first and last day's OQ scores to check for significant improvement overall as well as within each condition.

\subsubsection{Therapeutic outcomes via pre- and post-session anxiety scale scores}
To check for single-session improvements under each condition, we compared the participants' pre- and post-session anxiety scores on each day of the required week (days 1-8). We performed a dependent t-test (parametric) or Wilcoxon signed-rank test (non-parametric) between the pre-session and post-session anxiety scale scores on each day, depending on whether the data were normally distributed. To control the false discovery rate, we adjusted the alpha levels according to the Benjamini-Hochberg method~\citep{benjamini1995controlling}, applying the correction separately within each condition across the eight daily tests.

To account for the nested structure of the data (repeated observations within participants),  therapeutic outcomes were assessed using mixed-effects modeling. Linear mixed-effects models were estimated using the \texttt{lme4} package, with statistical significance evaluated using Satterthwaite’s approximation for degrees of freedom as implemented in the \texttt{lmerTest} package. An unconditional model was first estimated to calculate the intraclass correlation coefficient (ICC). Fixed effects included day in the study, condition (using the worksheet condition as the reference group), and both between-participant (grand-mean centered pre-session anxiety: pre\_gmc) and within-participant (cluster-mean centered pre-session anxiety: pre\_cmc) components of pre-session anxiety. This centering approach allowed for the disaggregation of within- and between-participant effects.
A bottom-up model building approach was used, where predictors were added sequentially. Nested models were compared using likelihood ratio tests to evaluate whether the inclusion of additional predictors significantly improved model fit.
The final model included random intercepts and random slopes for pre\_cmc across participants, allowing both baseline anxiety and within-participant associations to vary between individuals. Models were estimated using restricted maximum likelihood (REML). Missing data were handled using full information maximum likelihood (FIML), as implemented in \texttt{lme4}.

\subsubsection{Adherence to daily sessions} 
Adherence was calculated as the number of days in the second week of the study (days 9-15) that the participant completed a CBT homework session. To measure the difference in adherence to the CBT exercises between conditions, we performed a non-parametric version of one-way ANOVA (using Kruskal Wallis test) as the adherence data were not normally distributed but had homogeneous variance across conditions. To evaluate if participants' adherence was related to their attitudes towards robots (NARS), we fitted a simple linear regression model with adherence as the outcome. The assumptions of linear regression were met (linearity, independence of observations, normality of residuals (using the Shapiro-Wilk test), and homoscedasticity (using the Breusch-Pagan test)).

\subsubsection{Thematic Analysis of Exit Interviews}
Interview transcripts were analyzed in several iterations following thematic analysis methodology~\citep{clarke2014thematic}. 
During an initial familiarization stage, seven team members reviewed all interview transcripts, then met together to identify patterns in the data and develop a set of codes, following an inductive approach~\citep{clarke2014thematic}. 
After coding each transcript, the researchers reviewed the transcripts to develop and refine a thematic structure that reflected the participants' responses.

\subsubsection{LLM-Based CBT Exercise Analysis}

To determine how well the LLM-powered interactions aligned with the goals of the intended CBT exercise, two clinicians reviewed a randomly selected subsample (around 10\%) of the collected transcripts. Since CBT worksheets are quite varied, we did not use an explicit rubric for this evaluation; each of the clinicians relied on their extensive CBT experience to gauge if the exercise was being followed.  

%% file: sections/results.tex
\section{Results}\label{sec1}
\subsection{Baseline Assessments}
\subsubsection{Demographics}
Baseline demographics were measured through comparisons of the participants' age and gender (when reported) among the three conditions. Levene's test was not significant but as the data were not normally distributed, a non-parametric test was used to compare adherence among conditions. A Kruskal-Wallis test was performed on the scores of the three conditions. {The differences between the robot (mean = 20.07, rank mean = 16.82143), chatbot (mean =  20.75, rank mean = 20.83333), and worksheet (mean = 21, rank mean = 21.29167) conditions} were not significant, H(2, n = 38) =  1.3475, \emph{p} = 0.5098. To assess the distribution of gender across the three conditions, a Fisher’s exact test was used due to violations of chi-square assumptions (i.e., low expected cell counts). The test indicated no significant association between gender and condition, p = 0.351. These results show that there were no significant difference among the three conditions for key demographic data.

\subsubsection{Psychological Distress}

The data were normally distributed in each condition and passed the homogeneity of variance assumption test. A one-way ANOVA was performed on the scores of the Outcomes Questionnaire-45 (OQ) and the results indicated no significant differences across the three experimental conditions in their positive affect at baseline, F(2, 38) = 2.232, p = 0.1316. The estimated marginal mean scores were 67.4 for the robot condition (95\% CI [58.0, 76.7]), 53.5 for the chatbot condition (95\% CI [43.4, 63.6]), and 64.1 for the worksheet condition (95\% CI [53.5, 74.7]).

\subsubsection{Pre-Session Anxiety}
We also compared pre-session participant anxiety scores for the three conditions. Assumptions of normality and homogeneity of variance were satisfied; therefore, a one-way ANOVA was conducted. Significant between-condition differences were observed in baseline participant anxiety scores: F(2, 38) = 6.25, p = 0.007. Post hoc comparisons using Tukey’s HSD test indicated that anxiety scores were significantly lower in the chatbot condition (M = 46.8, 95\% CI [36.0, 57.6]) compared to the robot condition (M = 66.4, 95\% CI [56.7, 76.1]), $p = .025$. No other pairwise comparisons were significant ($p > .20$). The worksheet condition (M = 59.9, 95\% CI [49.1, 70.7]) did not significantly differ from the other two conditions ($p > .2$).


\subsection{Anxiety and Distress Measures}

\subsubsection{Statistical Comparison of Start to End of Study General Psychological Distress}
The Outcomes Questionnaire-45 (OQ) is a validated instrument designed to track weekly changes in mental health. It was selected for this experiment because it is commonly used to measure the effectiveness of psychotherapy. The OQ was used to assess self-reported measures of general psychological distress over the study period. The Cronbach's $\alpha$ showed high reliability for all days, $\alpha=0.89$ for Day 1 and $\alpha=0.93$ for Day 15. Dependent t-tests comparing Day 1 to Day 15 showed that there was a significant decrease in the OQ survey scores of participants across conditions at the end of the study (\emph{M}=52.73, \emph{SD}=20.068) compared to the beginning of the study (\emph{M}=61.892, \emph{SD}=17.810), t(36)=3.122, \emph{p} = 0.004, 95\% CI [3.262,15.062]. The effect size, as measured by Cohen's d, was $d$=0.513, a medium effect.

\begin{figure}[htbp]
    \centering
    \includegraphics[
        width=0.6\linewidth]{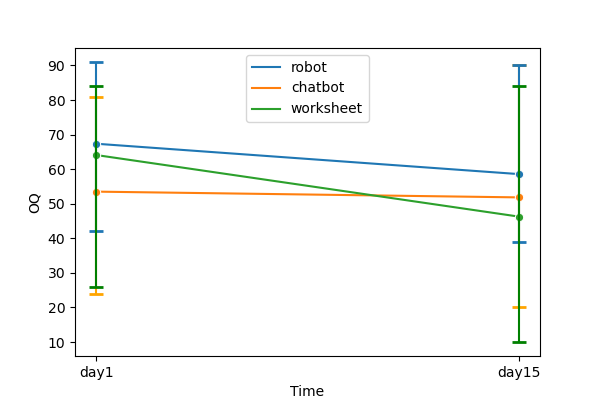}
    \caption{\textbf{Marginal means of OQ scores from the first to last day of the study.} There was a significant decrease in the OQ scores for the participants in the robot and worksheet conditions, but not in the chatbot condition.}
    \label{fig:OQ}
\end{figure}

There was also a significant decrease in the OQ survey scores of participants in the robot condition at the end of the study (\emph{M}=58.571, \emph{SD}=14.805) compared to the beginning of the study (\emph{M}=67.357, \emph{SD}=15.604), t(13) = 2.253, \emph{p} = 0.042, 95\% CI [0.367, 17.468]. {The effect size, as measured by Cohen's d, was $d= 0.602$, a medium effect.} 
There was no significant decrease in the chatbot condition (95\% CI [-11.307, 14.641]).
There was a significant decrease in the OQ survey scores of participants in the worksheet condition at the end of the study (\emph{M}=46.273, \emph{SD}=20.761) compared to the beginning of the study (\emph{M}=64.091, \emph{SD}=18.892), t(10)=4.046, \emph{p} = 0.002, 95\% CI [8.427, 29.095]. {The effect size, as measured by Cohen's d, was $d=1.22$, a large effect.}
Figure \ref{fig:OQ} shows Day 1 to Day 15 OQ score trends per condition.

\renewcommand{\figurename}{Fig.}

\subsubsection{Statistical Comparison of Pre-/Post-Session Anxiety Assessments}
The anxiety questionnaire (all $\alpha$'s $>$ 0.70) was administered before and after each CBT session, twice daily during the first week and, depending on participants' adherence, an average of 4-6 more times during Week 2. To determine how often a single CBT session was associated with significant anxiety reduction within each condition, we conducted paired-samples t-tests comparing daily mean pre-post session anxiety scores. Given the repeated day-specific comparisons, p-values were adjusted using the Benjamini–Hochberg procedure to control the false discovery rate (FDR) at \emph{q}=0.05. The FDR correction was applied separately within each condition across the eight daily tests. This procedure was selected because it controls the expected proportion of false discoveries among significant findings while preserving greater statistical power than family-wise error corrections. Accordingly, only day-specific effects that survived the Benjamini–Hochberg correction were interpreted as statistically significant.
Outcomes are summarized in Table \ref{tab:anxiety_change}, with detailed results in Table \ref{tab:anxiety_improvements}, which displays the days with significant pre- to post-session decreases and corresponding effect sizes.

Out of the eight mandatory days of CBT exercises, participants in the robot condition showed significant reductions in self-rated anxiety after Days 2, 3, 6, 7, and 8. Participants in the chatbot condition reported significant improvements on Days 1 and 7 (pre: \emph{Mdn} = 32.00; post: \emph{Mdn} = 30.00 at \emph{z} = -2.62, \emph{p} = .009, {$r = 0.826$}), and participants in the worksheet condition did not experience any significant anxiety reductions after any single session.

\begin{table}[htbp]
\centering
\caption{\textbf{Pre- to post-session anxiety score improvements by modality.}
For each study condition, the days where the change from pre- to post-session anxiety scores significantly improved are shown in green.}
\label{tab:anxiety_change}

\begin{adjustbox}{max width=\linewidth}

\arrayrulecolor{white}
\begin{tabular}{l|c|c|c|c|c|c|c|c}

& \multicolumn{8}{c}{Day} \\

Condition & 1 & 2 & 3 & 4 & 5 & 6 & 7 & 8 \\

Robot
& \cellcolor{nonsignificant}
& \cellcolor{significant}$\checkmark$
& \cellcolor{significant}$\checkmark$
& \cellcolor{nonsignificant}
& \cellcolor{nonsignificant}
& \cellcolor{significant}$\checkmark$
& \cellcolor{significant}$\checkmark$
& \cellcolor{significant}$\checkmark$ \\
\hline

Chatbot
& \cellcolor{significant}$\checkmark$
& \cellcolor{nonsignificant}
& \cellcolor{nonsignificant}
& \cellcolor{nonsignificant}
& \cellcolor{nonsignificant}
& \cellcolor{nonsignificant}
& \cellcolor{significant}$\checkmark$
& \cellcolor{nonsignificant} \\
\hline

Worksheet
& \cellcolor{nonsignificant}
& \cellcolor{nonsignificant}
& \cellcolor{nonsignificant}
& \cellcolor{nonsignificant}
& \cellcolor{nonsignificant}
& \cellcolor{nonsignificant}
& \cellcolor{nonsignificant}
& \cellcolor{nonsignificant} \\

\end{tabular}
\arrayrulecolor{black}

\end{adjustbox}
\end{table}

\begin{table}[h!]
\centering
\caption{\textbf{Pre- to post-session anxiety improvements.} Only days with significant reductions in anxiety ($p < .05$) are shown. The corrected p values after the Benjamini-Hochberg procedure are given in column p-BH. The results for Chatbot Day 7 were computed using Wilcoxon signed-rank test because the assumption of homogeneity of variance was violated. }

\begin{adjustbox}{max width=\linewidth}
\begin{tabular}{c c c c c c c c c}
\hline
Condition & Day & Pre-Session M (SD) & Post-Session M (SD) & Test Statistic & p-value & p-BH & Cohen's d & 95\% CI \\
\hline
Robot & 2 & 64.68 (18.93) & 58.75 (21.30) & t(13) = -2.981 & 0.011 & 0.024 & 0.797 & [-21.474, -3.428] \\
Robot & 3 & 61.93 (24.20) & 51.00 (11.99) & t(13) = -2.795 & 0.015 & 0.024 & 0.747 & [-17.597, -2.252] \\
Robot & 6 & 55.08 (17.74) & 47.62 (15.56) & t(12) = -3.757 & 0.003 & 0.022 & 1.04 & [-13.202, -3.510] \\
Robot & 7 & 55.79 (21.85) & 46.86 (16.13) & t(13) = -3.282 & 0.006 & 0.024 & 0.877 & [-14.806, -3.051] \\
Robot & 8 & 60.64 (29.40) & 51.00 (20.10) & t(13) = -2.838 & 0.014 & 0.024 & 0.758 & [-13.196,  -1.788]\\
Chatbot & 1 & 46.83 (12.38) & 44.75 (18.21) & t(11) = -3.867 & 0.003 & 0.021 & 1.12 & [-9.344, -2.565] \\
Chatbot & 7 & 38.92 (14.16) & 35.00 (11.92) & z = -2.620 & 0.009 & 0.035 & 0.826 & [-11.000,  -1.500] \\
\hline
\end{tabular}
\end{adjustbox}
\label{tab:anxiety_improvements}
\end{table}

The data had a two-level structure, with repeated observations (Level 1) nested within participants (Level 2). The sample included 38 participants contributing a total of 297 observations, with an average of approximately 8 observations per participant.
An unconditional (intercept-only) model indicated that 65.2\% of the variance in post-session anxiety was attributable to between-participant differences (ICC = .65), suggesting substantial clustering of observations within participants and supporting the use of a mixed-effects modeling approach.
A linear mixed-effects model was conducted to examine whether post-session anxiety was predicted by condition (worksheet, robot, and chatbot), day in the study, and both between-participant (grand-mean centered pre-session anxiety; pre\_gmc) and within-participant (cluster-mean centered pre-session anxiety; pre\_cmc) components of pre-session anxiety. Random intercepts and random slopes for pre\_cmc were included for participants.
There was a significant main effect of condition. Compared to the worksheet condition, participants in the robot condition reported significantly lower post-session anxiety (b = -6.26), (p = .001), as did participants in the chatbot condition, (b = -6.74), (p = .002). The effect of day was not significant, (b = -0.33), (p = .111).
There was a significant positive effect of between-participant pre-session anxiety, such that individuals with higher average pre-session anxiety reported higher post-session anxiety, (b = 1.13)(p $<$ .001). Similarly, within-participant fluctuations in pre-session anxiety were positively associated with post-session anxiety, (b = 0.63)(p $<$ .001).
Condition significantly moderated the between-participant effect of pre-session anxiety. The association between pre\_gmc and post-session anxiety was lower in both the robot condition, (b = -0.44),(p $<$ .001), and the chatbot condition, (b = -0.41), (p = .003), relative to the worksheet condition. However, the condition did not significantly moderate the within-participant effect of pre-session anxiety (p $>$ .05).
Random effects indicated substantial variability in baseline post-session anxiety across participants (($\sigma^2 = 11.64$), (SD = 3.41)), as well as variability in the within-participant effect of pre-session anxiety (($\sigma^2 = 0.10$), (SD = 0.32)). The residual variance was 55.23 ((SD = 7.43)). These results are further detailed in Table \ref{MLM_tabel}, and the model equations are outlined below.

\begin{table}[ht]
\centering
\caption{\textbf{Linear Mixed-Effects Model Results Predicting Post-Session Anxiety.} The worksheet condition was taken as the base model. 297 observations nested within 38 participants. pre\_gmc = rand-mean centered pre-session anxiety; pre\_cmc = cluster-mean centered pre-session anxiety. \\ Significance codes: *** 0.001 ** 0.01 * 0.05 . 0.1 ' ' 1}

\begin{tabular}{rrrrrrl}
  \hline
 & Estimate & Std. Error & df & t value & Pr($>$$|$t$|$) &  \\ 
  \hline
(Intercept) & 53.16 & 1.45 & 56.89 & 36.73 & $<$ 2e-16 & *** \\ 
robot & -6.26 & 1.80 & 32.49 & -3.48 & 0.0014 & ** \\ 
chatbot & -6.74 & 2.04 & 33.01 & -3.31 & 0.0023 & ** \\ 
day\_number & -0.33 & 0.21 & 255.50 & -1.60 &  0.1112 &  \\ 
pre\_gmc & 1.13 & 0.08 & 33.49 & 14.80 & 2.98e-16 & ***  \\ 
pre\_cmc & 0.63 & 0.14 & 40.54 & 4.63 & 3.74e-05 & *** \\ 
robot:pre\_gmc & -0.44 & 0.10 & 32.45 & -4.35 & 0.0001 & ***  \\ 
chatbot:pre\_gmc & -0.41 & 0.13 & 33.35 & -3.17 & 0.0033 & ** \\ 
robot:pre\_cmc & -0.11 & 0.17 & 34.97 & -0.65 & 0.5180 & \\ 
chatbot:pre\_cmc & -0.21 & 0.20 & 40.02 & -1.09 & 0.2833 & \\ 
\hline
\multicolumn{6}{l}{\textit{Random effects}} \\ 
\hline
participant\_id (Intercept) & 11.64 & 3.41 & & & \\ 
participant\_id (pre\_cmc) & 0.10 & 0.32 & & & \\ 
Correlation (Intercept, pre\_cmc) & 0.47 & & & & \\ 
Residual & 55.23 & 7.43 & & & \\ 
\hline
\multicolumn{6}{l}{\textit{Observations: 297, Groups: 38}} \\
\hline
\end{tabular}
\label{MLM_tabel}
\end{table}


The following equations detail the mathematical framework of the linear mixed-effects model, specifying how post-session anxiety is determined by the experimental condition, study duration, and the partitioned components of pre-session anxiety.

Level 1: Within-Participant
\begin{align*}
\text{post}_{j} &= \beta_{0j} 
+ \beta_{1}(\text{robot}_{j})
+ \beta_{2}(\text{chatbot}_{j})
+ \beta_{3}(\text{day}_{j}) \\
&\quad + \beta_{4}(\text{pre\_gmc}_{j})
+ \beta_{\text{pre\_cmc},j}(\text{pre\_cmc}_{j}) \\
&\quad + \beta_{6}(\text{robot}_{j} \times \text{pre\_gmc}_{j})
+ \beta_{7}(\text{chatbot}_{j} \times \text{pre\_gmc}_{j}) \\
&\quad + \beta_{8}(\text{robot}_{j} \times \text{pre\_cmc}_{j})
+ \beta_{9}(\text{chatbot}_{j} \times \text{pre\_cmc}_{j})
+ r_{j}
\end{align*}

Level 2: Between-Participant
\begin{align*}
\beta_{0j} &= \gamma_{00} + u_{0j} \\
\beta_{\text{pre\_cmc},j} &= \gamma_{50} + u_{1j}
\end{align*}

Combined model (with fixed effects coefficients):
\begin{align*}
\text{post}_{j} &= 53.16 
- 6.26(\text{robot}_{j})
- 6.74(\text{chatbot}_{j})
- 0.33(\text{day}_{j}) \\
&\quad + 1.13(\text{pre\_gmc}_{j})
+ 0.63(\text{pre\_cmc}_{j}) \\
&\quad - 0.44(\text{robot}_{j} \times \text{pre\_gmc}_{j})
- 0.41(\text{chatbot}_{j} \times \text{pre\_gmc}_{j}) \\
&\quad - 0.11(\text{robot}_{j} \times \text{pre\_cmc}_{j})
- 0.21(\text{chatbot}_{j} \times \text{pre\_cmc}_{j}) \\
&\quad + u_{0j} + u_{1j}(\text{pre\_cmc}_{j}) + r_{j}
\end{align*}

where $i$ is an observation from participant $j$, $u_{0j}$ is the random intercept for participant $j$, $u_{1j}$ is the random slope for pre\_cmc for participant $j$ , and $r_{j}$ is the residual.

By combining these levels, the final equation accounts for the specific predictive power of the robot and chatbot interventions while simultaneously controlling for random error and participant-level variability.

To evaluate the sensitivity of the mixed-effects model, simulation-based power analyses were conducted using the simr package with 1,000 simulations. Power was estimated for the primary fixed effects of interest. The model demonstrated high power to detect the observed effect of condition (96.4\%, 95\% CI [95.1\%, 97.5\%]) and the observed condition × baseline anxiety interactions (97.0\%, 95\% CI [95.7\%, 98.0\%]). In contrast, power to detect the observed effect of study day was comparatively low (38.7\%, 95\% CI [35.7\%, 41.8\%]), suggesting that effects of this magnitude may have been difficult to detect with the current sample size n = 38. Because the observed power for the day-level effect was below conventional thresholds, we conducted a simulation-based minimum detectable effect size analysis using the simr package. Results indicated that, given the current sample size and model structure, an effect of approximately $\beta$ = 0.61 would be required to achieve 80\% power for detecting the day-level fixed effect.

\subsection{Usability}

\subsubsection{Statistical Comparison of SUS}
A one-way ANOVA test was performed on the system usability scale (SUS) scores of the three conditions. The differences between the robot (mean = 76.250, rank mean = 20.786), chatbot (mean = 75.833, rank mean = 20.958) and worksheet (mean = 68.958, rank mean = 16.542) conditions were not significant, H(2, n = 38) = 0.992, \emph{p} = 0.381, {$ \eta^2 = 0.054$, therefore, there was no effect.}

\subsubsection{Statistical Comparison of Session Duration}

A Kruskal–Wallis test was conducted to examine differences in session duration across conditions. There was no statistically significant difference between groups, H(2, N = 38) = 2.33, p = .312. Median session durations were 6.58 minutes (95\% CI [3.89, 10.59]) for the robot condition, 4.79 minutes (95\% CI [2.69, 5.88]) for the chatbot condition, and 5.57 minutes (95\% CI [3.77, 7.50]) for the worksheet condition. The overlapping confidence intervals suggest comparable session durations across conditions.

\subsection{Adherence}
\subsubsection{Statistical Comparison of Days of Optional Use}
\begin{figure}[h]
\centering
\includegraphics[width=0.6\linewidth]{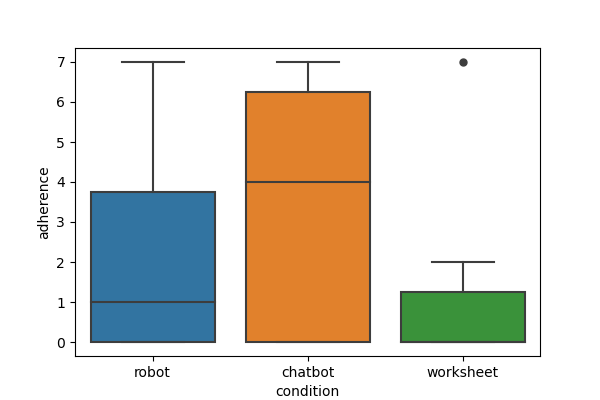}
\caption{\textbf{Adherence means.} The distribution of adherence, measured by the number of days participants completed the exercises during the optional week (Days 9-15 of the study), for each condition.}
\label{fig:H3Boxplot}
\end{figure}

Adherence was measured as the number of days in Week 2 that participants continued to use their system for CBT exercises. Levene's test was not significant; as the data were not normally distributed, a non-parametric test was used to compare adherence between conditions. A Kruskal-Wallis test was performed on the scores of the three conditions. {The differences between the robot (mean = 2.07, rank mean = 19.46), chatbot (mean = 3.67, rank mean = 23.96), and worksheet (mean = 1.08, rank mean = 15.08) conditions} were not significant, H(2, n = 38) = 4.2, \emph{p} = 0.1, $\eta^2 = 0.062$ (See Figure \ref{fig:H3Boxplot}).

\subsubsection{Correlation between Attitude towards Robots and Adherence}
A simple linear regression indicated that there was a significant effect between the adherence and a participant's negative attitude towards robots (NARS) (\emph{F(1, 9)} = 9.139, \emph{p} = 0.014, $R^2$ = 0.449). The independent variable NARS was examined further and was a significant predictor of adherence (\emph{t} = -3.02, \emph{b} = -0.339, \emph{p} = 0.014, 95\% CI [-0.593, -0.085]).

\subsection{LLM-Based CBT Exercise Analysis}

Two clinicians evaluated around 10\% of the LLM-based condition transcripts. They found that the LLM initiated exercises aligned closely with the exercise structure (for both cognitive restructuring and coping strategies) at the beginning of interactions and was able to incorporate a variety of CBT skills and techniques beyond the worksheet’s requirements, such as including psychoeducation to facilitate participant insight and understanding.
After beginning an exercise, the LLM demonstrated active collaboration with the participant and could appropriately shift away from the worksheet structure to provide needed psychoeducation or discuss barriers to treatment that were preventing the participant from engaging in the designated, structured CBT technique. These findings are reflective of similar traditional in-person psychotherapy practices that focus on building rapport and encouraging treatment “buy-in” to facilitate stronger treatment adherence \cite{leach2005rapport}. 

We also compared the conversation lengths in the chatbot condition and the robot condition. The results of a Welch’s two-sample t-test indicated that there was no significant difference in the average number of turns between the chatbot condition (M = 11.28) and the robot condition (M = 12.01), t(275.78) = -0.93, p = .35, 95\% CI [-2.27, 0.81]. A second  Welch’s two-sample t-test indicated that the average word count did not differ significantly between the chatbot condition (M = 168.56) and the robot condition (M = 225.23), t(23.01) = -1.03, p = .31, 95\% CI [-170.13, 56.80].


\subsection{Thematic Analysis}

\subsubsection{CBT Delivery Method}

\paragraph{Worksheet}

Participants completing CBT worksheets generally found the web-based worksheet interface easy to use, with little to no learning required. 
They reported that completing daily worksheets led them to reserve time each day for reflection. However, many participants found that static questions did not always align with their specific problems and wanted more guidance or dynamic follow-up questions.
For example, P38 stated: \textit{``It is a little one-sided: I am just writing what I am feeling I am not able to discuss with anyone. I am not able to get an alternate version of what I am feeling.''}

\paragraph{Chatbot}
In the chatbot condition, participants found the chatbot useful and simpler to use than they expected. 
Many found it engaging, and some participants even used it beyond the required eight days during the remainder of the two-week study period. 
Like those in the worksheet condition, some participants enjoyed writing out their thoughts. 
They appreciated the informal, conversational style, comparing it to texting, and valued the opportunity to write out their thoughts.
However, many participants reported a lack of social connection with the chatbot, feeling that, as P16 stated, \textit{``There is nobody there. It's just a bot.''} Other participants commented that the chatbot could be helpful for others, but not for them. 

\paragraph{Robot}
Participants in the robot condition reported feeling nervous on the first day. They were concerned about using it incorrectly and somewhat hesitant to interact with the robot, which was an unfamiliar process for them. P4 recounted, \textit{``I was a little bit hesitant at first. It was a little interesting to be talking to a robot"}. After the first session, participants in the robot condition generally reflected positively about their interactions, reporting that completing exercises with a physical robot was fun and engaging.
Blossom's zoomorphic, non-threatening design was often described as cute, approachable, and supportive. For example, P10 affectionately saying \textit{``It is like a friend or a pet.''} 
Many participants appreciated the robot's movements and deep breathing, which made it feel more lifelike, though a few initially found these movements uncomfortable. 
Some expressed a desire to customize the robot’s appearance or voice.  P3 was \textit{``shocked to hear Blossom's deep voice, [...] expect[ing] a cartoon character voice''} instead. 
Overall, participants emphasized that the robot’s physical presence and verbal output enhanced the experience and made the CBT exercises feel more human. P5 said: \textit{"It was essential to have the robot rather than just the computer system [...] when it's just [chatbot messages], it doesn't feel as human as it would be with a robot.''}

\subsubsection{LLM Performance}

Both the chatbot and the robot were powered by OpenAI's GPT-3.5 LLM. We evaluated how participants perceived the quality, engagement, and usefulness of their conversations to account for the fact that LLMs can sometimes produce inconsistent or unexpected responses.

\paragraph{Chatbot}

Participants in the chatbot condition were generally impressed by its conversational abilities, appreciating the personalized responses and noting that it felt surprisingly human-like and flexible. P16 even said, \textit{``At first questioning if I was actually talking to AI. It felt like a person.''} 

However, several challenges emerged. 
Some participants felt the chatbot offered generic advice or lacked empathy, and a few wanted it to take a leading role in suggesting coping strategies.
Two participants reported feeling disheartened because they perceived that the chatbot did not care about what they were saying.
Participants sometimes had to work to sustain conversations, and the chatbot occasionally ended sessions prematurely. P18 complained that 
\textit{``Sometimes it does cut you off at the end.''} 
While the chatbot did not retain any information between sessions, occasional hallucinations led some participants to believe it remembered prior conversations, and many expressed a desire for it to do so. Over time, participants learned that providing the chatbot with more details elicited richer, more relevant responses.

\paragraph{Robot}

Similarly, participants in the robot condition were impressed with the robot's conversational abilities, but encountered a couple of challenges. They sometimes had to take the lead in carrying the conversation and were frustrated when the robot tried to end the dialogue prematurely. Participants also wanted deeper discussions and more guidance beyond the robot’s generic suggestions. 
As P1 expressed it, \textit{``It was the same responses every time, it always suggested to calm down.''}.
As with the chatbot, the robot occasionally implied memory of prior sessions or planning for future ones (e.g., ``Let's talk about this next time''), leading some participants to believe it could remember past conversations. Many suggested improving the robot by enabling it to reliably recall previous sessions to support continuity and personalized guidance.

\subsubsection{Therapeutic Experience and Outcomes}

The study was intentionally conducted during the last two weeks of the academic semester, a period of naturally heightened stress for students. In this section, we discuss the participants’ experiences with the CBT exercises across the three delivery methods, examining how they engaged with the activities, reflected on their stressors, and applied the strategies in their daily lives.

\paragraph{Worksheets}

Participants in the worksheet condition emphasized that completing the daily exercises encouraged self-reflection, a core component of CBT~\citep{beck2020cognitive}. Many noted that writing about their thoughts helped them to identify illogical patterns of thinking and to explore solutions to their challenges.
P36 shared: \textit{``It was beneficial to see exactly what was just written. If I saw something that was not a logical thought that I had just written, I could catch myself.''}
Several participants reported that the exercises reduced stress and helped them to apply coping strategies in real time.  P32 said:
\textit{``One time, I was studying for a test I had to do and I was very frustrated because I didn't get much time. I got reminded about the survey and I did it right there and then. I got through a problem-solving [exercise] [...] and realized I could use the strategies.'' }

However, some participants felt that reflecting on their problems did not relieve stress or address long-term challenges. P37 noted: \textit{``I guess I felt a bit better talking about it, but it didn't relieve the stress because it didn't do my final or assignments for me [...] I kind of felt stressed afterwards because I know it's still a problem.''}
A few participants who already considered themselves analytical or self-aware found the worksheets less helpful. P36 suggested: 
\textit{``CBT is good especially for people who can't see through their own thoughts, but for people who are self-aware and able to see things from different points of view and take them to heart, CBT might not be the way to go. They might need more of a solutions-based approach.''}

\paragraph{Chatbot}

Participants in the chatbot condition generally found the CBT exercises effective and appreciated talking through their stressors, with P23 reporting, \textit{``After the end of the session I was completely relieved and less stressed.''} 
Similarly, many participants stated that using the chatbot decreased their stress by helping them explore their feelings and find a more positive outlook.  P18 reported: \textit{``It's not the end of the world; it was helpful to remind myself of that sometimes.''} 
Participants familiar with CBT reported that the exercises met their expectations, while newcomers felt the chatbot provided a helpful introduction. Some participants reported improved goal-setting, better reframing of negative experiences, and increased openness to seeking therapy.

Although most participants appreciated the interactions, some found the chatbot frustrating, reporting that it focused too heavily on negative emotions, offered overly generic solutions, or did not provide sufficient empathy.
P21 recalled: \textit{``The chatbot was too abrupt and very focused on just negative emotions, just trying to get the negative emotions out of me and nothing else.''} One participant similarly felt that the chatbot could not provide much emotional gratification and lacked a sense of understanding.   
Others noted that exercises sometimes increased stress by prompting reflection on negative experiences. For example, P22 said: \textit{``I would leave feeling more upset because I would be thinking about more negative thoughts [...] issues I wasn't already thinking about.''} A few participants also reported that the exercises had little effect because they did not apply the coping strategies or internalize the process.

\paragraph{Robot}

Participants in the robot condition overall reported a positive experience, often noting reduced stress and anxiety.  For example, P5 reported: \textit{``I had fewer panic attacks in general. If I was feeling really anxious I could sit down with Blossom, and it overall reduced my anxious behavior.''}
The robot facilitated reflection on daily stressors and helped participants restructure thoughts and identify solutions. 
Participants new to CBT found it to be an accessible introduction, with P6 noting: \textit{``[CBT] feels a lot less intimidating to have as a resource and it has less stigma in [their] mind''}, and other participants echoing this sentiment. 
Many participants felt the robot validated their experiences, provided reassurance, and offered empathetic support. They reported applying coping strategies learned with the robot in everyday situations and becoming more socially engaged, such as P14, who stated:
\textit{Blossom made me feel less socially anxious''}. P4 noted: \textit{One thing Blossom said was `go and talk to people who really care about you,' so then I went to my best friend and it helped me address problems I have been putting off.''} 

However, some participants reported similar frustrations as those in the chatbot condition, including decreased mood after reflecting on stressors and negative thoughts, insufficient comfort, or perceived disregard for their problems. One participant also pointed out that a human therapist could provide more nuance.

\subsubsection{Motivation and Frequency of Use}

Participants’ engagement with the exercises varied across the three delivery methods, influenced by both personal preferences and external factors such as time constraints. Next, we examine how often participants chose to complete the exercises, their motivations, and factors that encouraged or prevented use in each condition.

\paragraph{Worksheet}

Participants had mixed preferences for the frequency of worksheet completion. Many preferred to use the worksheets on an as-needed basis, stating that some days they had no pressing issues to reflect on.  According to P30: 
\textit{``Some days there was no need, I was doing them only when I needed to structure my thoughts.''}
Those who found the worksheets helpful in addressing daily challenges tended to complete them daily, and some even developed a habit of using them consistently, with P39 noting, \textit{``It became a habit for me to sit down and do it [...] I'm pretty sure I'll be using this for a long time.''}  

Many participants did not continue to complete worksheets into the second week. Motivation waned for several participants due to external factors such as busy schedules, difficulty finding a private space, and the effort required to complete the exercises. Some participants also noted a lack of accountability, including the fact that there were no measures in place to keep them from speeding through the activity.

\paragraph{Chatbot}

Similar to the worksheet condition, many participants reported using the chatbot on an as-needed basis, particularly when they experienced stress. Some ran out of topics to discuss, while others continued because new issues arose. Time constraints were a common reason for reduced engagement in the second week. P28 stated: 
\textit{``I did not continue with the exercises, mostly because as a college student, I don't have time to do what I want to do. I'm a biomedical engineering major. That takes up a lot of my time.''} 
Novelty and curiosity about the chatbot’s responses encouraged some participants to continue, while others suggested adding accountability mechanisms like follow-ups with a human therapist to increase motivation.

\paragraph{Robot}

Participants in the robot condition reflected more positively on completing daily sessions than participants in the chatbot and worksheet conditions, observing that they felt better with consistent use. Participants were generally excited and curious to interact with the robot, finding it novel and engaging. Many reported that daily interactions helped them establish a routine.  For example, P6 stated: \textit{``Continuing CBT with this system really encourages you to do it every day.''} Still, some participants reported that they only used the robot when stressful events had occurred, such as P7, who said: \textit{``When I feel crappy that day, I go to Blossom. It was a good resource for me''}. Only two participants reported wanting less frequent interactions with the robot. 

The physical presence of the robot served as a reminder to complete exercises, though some participants noted that it lacked the authority of a human therapist. P8 said: 
\textit{``A robot that cute doesn't give as much authority as a licensed therapist.''} During the second week, many participants continued using the robot regularly, and some found that repeated interactions allowed for deeper reflection. A few participants adjusted the duration or frequency to suit their schedules, emphasizing the flexibility of the robot as an advantage. Participants who decreased or discontinued the use of the robot in the second week often cited external reasons, such as being busy with schoolwork.

%% file: sections/discussion.tex
\section{Discussion}\label{sec1}

This work leveraged SAR and LLM technologies to support CBT homework exercises. 
We compared three interfaces for facilitating regular at-home CBT homework exercises for college students.
Most interestingly, we found a paradox where the least flexible interface, the worksheet condition, led to the greatest longitudinal improvement in therapeutic outcomes, while the interactive LLM-based conditions had the strongest pre- to post-session anxiety reductions. We ground our discussion of the results in both the quantitative and qualitative findings of this exploratory study.

\subsection{Therapeutic Outcomes}

\subsubsection{Weekly Therapeutic Outcomes}
To address \textbf{RQ1}, which examined how the CBT in-home exercise delivery method affects short-term therapeutic outcomes, we compared students' Day 1 OQ scores with their Day 15 scores, because this validated instrument tracks weekly changes in mental health and measures the effectiveness of psychotherapy.
A dependent t-test found a statistically significant decrease in OQ across all conditions, suggesting that sustained engagement with CBT exercises over a two-week period is effective for decreasing psychological distress. 

A condition-by-condition analysis of changes in OQ score by the end of the intervention provided interesting insights. 
The worksheet condition showed the largest effect on the students' differences in OQ scores, followed by the robot condition, and there was no statistically significant improvement found in the chatbot condition. Unsurprisingly, the worksheet condition produced large improvement in OQ scores, consistent with the extensive clinical use and empirical support for worksheet-based CBT interventions. 
The robot condition also produced statistically significant improvements, suggesting that SAR-delivered CBT is a viable modality, possibly because of an increased sense of participant accountability.
In contrast, it was unexpected that the chatbot condition had no significant difference in OQ scores, as existing research has suggested that chatbot-guided therapy can lead to positive therapeutic outcomes~\citep{jang2021mobile}. The chatbot appeared to suffer from the ``worst of both worlds", lacking both the clinical validation of worksheets and the physical presence that made the robot engaging.

Since there was no statistically significant difference in Day 1 OQ scores or demographic makeup between the condition groups, it is unlikely that baseline group differences could account for the varied outcomes. However, there could be other baseline differences between groups that we failed to capture; for example, the time of day at which exercises were most frequently completed. Furthermore, because we did not have a content moderator to ensure that the robot and chatbot conditions stayed directly aligned with the worksheet content, we could have had differences in CBT content delivery across conditions, which would have influenced participant outcomes. Similarly, we had no formal measure of how deeply participants engaged with exercises or how motivated they felt, which could be confounding variables in achieving therapeutic outcomes. 
Overall, these results should be interpreted with caution, given the limited sample size across conditions. Nonetheless, these trends provide an intriguing look into the effectiveness of SAR-assisted CBT exercises. 

\subsubsection{Single-Session Therapeutic Outcomes}
While the previous analysis examined short-term therapeutic outcomes across the two-week intervention, \textbf{RQ1} also considers immediate therapeutic outcomes. To address this aspect of the research question, we analyzed anxiety ratings at the beginning and at the end of each session.

As a baseline assessment, we examined whether pre-session anxiety scores on Day 1 differed across the three conditions. Results showed that participants in the chatbot condition reported significantly lower pre-session anxiety than those in the robot condition, while no significant differences were observed between the worksheet condition and either of the other two conditions.
These findings are consistent with post-study interview feedback, in which more participants in the robot condition expressed feeling anticipatory nervousness about the upcoming interaction on Day 1 than participants in the other conditions. Given that this instrument was measured at the start of the first session (once the robot was set up and running on the participant’s desk), we interpret this difference as part of settling into the study and getting used to using the robot. Participants in the other conditions reported greater familiarity and preparedness, drawing on prior experiences with worksheets or chatbots.

We calculated whether there was a decrease in participant anxiety from the start to the end of each mandatory session for each condition. Interestingly, we found that the worksheet condition showed no statistically significant immediate reductions in anxiety, the chatbot condition showed a statistically significant improvement in participants' anxiety on two of the eight mandatory days, and the robot condition showed improvement on five of the eight days.

The lack of immediate anxiety reduction in the worksheet condition may have been because the worksheet exercises were more cognitively demanding and less exciting to complete than engaging with an LLM-based interactive system. However, participants still experienced long-term improvements in the worksheet condition, demonstrating that though individual sessions did not provide an improvement pre-/post-session, they enabled participants to internally process their emotions and learn CBT skills for sustained change, consistent with the literature on the value of CBT homework exercises \citep{kazantzis2010meta}.

The chatbot modality had some immediate effects, but lacked significant long-term distress reduction. This highlights an important distinction between exercises that may feel satisfying or effective in the moment, but may not translate into long-term therapeutic benefit.

The robot condition demonstrated the most frequent statistically significant improvements, with over half of the mandatory sessions yielding a reduction in anxiety. This suggests two possibilities: 1) the robot-guided interaction was inherently more effective for immediate anxiety reduction; or 2) the robot-guided interaction caused the participants to start each session in a more elevated, anxious state that then was reduced through the CBT exercise. These insights merit further in-depth exploration, given that the robot condition was the only modality to have both session-level and long-term improvements.

Due to the hierarchical nature of our data, in addition to the single session anxiety assessment, we developed a mixed-effects model to assess the pre- to post-session anxiety scores.
In our mixed-effects model, the significant main effect of study condition suggests that both the robot and chatbot conditions were associated with lower post-session anxiety compared to the worksheet condition. On average, participants in these conditions ended sessions with anxiety scores approximately 6–7 points lower than those in the worksheet condition, when controlling for other variables. There was no strong evidence of a systematic change in anxiety over time.

Pre-session anxiety was a strong predictor of post-session anxiety at both the between-participant and within-participant levels. Individuals with higher-than-average pre-session anxiety (relative to others) had post-session anxiety scores that were, on average, 1.13 points higher for each unit increase in pre-session anxiety. Additionally, on days when a participant’s pre-session anxiety was higher than their own personal average, their post-session anxiety was also higher by approximately 0.63 points. This indicates that between-participant differences in anxiety are more strongly associated with outcomes than within-participant fluctuations.
Furthermore, individuals with higher pre-session anxiety not only reported higher post-session anxiety overall, but also exhibited greater reactivity to fluctuations in their anxiety. In other words, deviations from their typical anxiety levels had a stronger impact on their post-session outcomes. This pattern is consistent with the idea that individuals with elevated anxiety may have more difficulty regulating momentary increases in distress, making therapeutic engagement more challenging.

Interestingly, there was a significant interaction between condition and average pre-session anxiety (between-participant effect). Although higher baseline anxiety was associated with higher post-session anxiety in general, this relationship was curbed in both the robot and chatbot conditions. Specifically, the increase in post-session anxiety associated with higher pre-session anxiety was reduced by approximately 0.4 points per unit increase in pre-session anxiety in these conditions. This suggests that the robot and chatbot interventions were particularly effective for individuals with higher-than-average anxiety. In contrast, for individuals with lower-than-average pre-session anxiety, the worksheet condition showed relatively better outcomes. Taken together, these findings suggest that the robot and chatbot interventions may be especially beneficial for individuals with elevated anxiety, while traditional approaches may be sufficient or even preferable for those with lower baseline anxiety. Given that CBT is often used to support individuals with various anxiety disorders, observed effectiveness of the robot and chatbot interaction among individuals with elevated anxiety is promising. However, further research is needed to better understand these effects.

As with the weekly therapeutic outcomes, there may be additional factors that contributed to changes in single-session therapeutic outcomes by condition. For example, participant motivation and session timing throughout the day were not captured in the current study, leaving opportunities for future research.

\subsection{Adherence Outcomes}
To address \textbf{RQ2}, we examined participants' adherence to daily CBT exercises across the three delivery modalities. There was no statistically significant difference in adherence to daily CBT exercises across the three conditions. However, participants' post-study interviews provided insight into their experiences with the different modalities and the factors that influenced their willingness to continue engaging with CBT exercises. While these qualitative findings do not indicate differences in actual adherence behavior, they reveal that participants perceived the three modalities differently and cited different motivations and barriers related to continued use of each one after the required period.

Across all three conditions, participants were instructed to complete CBT exercises daily. In each modality, participants suggested that the exercises might be more helpful at a reduced frequency or on an as-needed basis, an approach several participants enacted in the second week of the study, when practice was optional. In all conditions, some participants reported running out of topics to discuss, which may have led them to complete fewer sessions in the second week. Some participants also reported wanting to take time to practice applying the skills they learned from the exercises in their daily lives before completing additional sessions, decreasing daily participation. These themes suggest that daily CBT exercises may exceed a reasonable level of interaction for most participants. 

Logistical constraints were another commonly reported reason for decreased use of the CBT exercise systems. For instance, the timing of the study partially overlapped with some participants' final exams,  limiting their capacity to engage in the daily CBT exercises. Some participants also fell ill, negatively impacting their adherence. Additionally, the sessions had to be completed in a private setting due to IRB restrictions, and some participants found it challenging to find a time to be alone to complete sessions. These common logistical barriers may partly explain why adherence did not differ significantly between modalities, since participants faced similar practical constraints regardless of which system they used.

Interestingly, participants only cited the time-consuming nature of the pre- and post-surveys and the lack of accountability as deterring them from completing additional sessions in the worksheet and chatbot conditions. This may suggest that, in the robot condition, participants felt a greater sense of social obligation toward the physical robot than they did toward the other modalities, even if they did not act on it.

In the two LLM conditions (robot and chatbot), participants said that the ability to complete flexible-length sessions made the system easier to fit into their busy schedules, an advantage of these modalities suggesting that even though adherence rates were comparable across conditions, participants in the LLM conditions may have engaged more willingly and with greater enjoyment.

In the robot conditions, participants also reported that the robot's physical presence on their desk served as a visual reminder to complete the exercises. Even though this did not produce measurable increases in adherence, it points out the potential of embodiment to serve as a passive reminder toward promoting adherence. 

Overall, although adherence trends for optional engagement with CBT exercises did not vary significantly by modality, the participants' perspectives on their interactions did, and users chose to speak about varying reasons that either encouraged or discouraged them from using the platforms. Another important insight from participants was that lower daily adherence did not inherently indicate a lack of interest, but may in some cases have reflected deeper engagement with the material.

\subsection{User Preferences}

\subsubsection{System Usability}
To address \textbf{RQ3a}, we examined participants' perceptions of the platform structure through usability ratings.
High usability ratings across all conditions ($M > 68$) suggest that the participants found each CBT system to be simple to use. The lack of significant differences in usability across the three conditions indicates that the robot's physical complexity did not make it more difficult for participants to use the platform than the  worksheet.

\subsubsection{LLM-Based Exercises}
To address \textbf{RQ3b}, we examined participants' perceptions and experiences with the LLM-driven CBT exercises through post-study interviews.
Our post-study interviews highlighted that participants in the two LLM-based conditions enjoyed the flexibility of the dialog, while the participants in the worksheet condition sometimes felt limited by the static format. Participants in the worksheet condition mentioned more guidance for the worksheets would be helpful, complained that the questions felt too generic and sometimes did not align with the specific topic they wanted to discuss, and compared the system to journaling instead of to talk therapy. Participants in the LLM-based conditions, on the other hand, expressed that the interaction was fun and engaging, they were impressed with the flexible dialogue, and compared the systems to human therapists.

The human-like nature of the conversational LLM raised participants' expectations of the system, which, at times, led to subsequent disappointment and frustration when the LLM could not match the quality of a human therapist. Despite these occasional letdowns, participants appreciated the judgment-free nature of the LLM system, emphasizing that they felt comfortable being honest with a chatbot or robot without fear of being judged.

Overall, participants in the robot and chatbot exercise conditions reported positive experiences, indicating that LLMs can enable engaging CBT homework exercises. They present a viable and novel alternative to traditional CBT worksheets and can expand the set of tools available to therapists in their practice.

\subsubsection{Robot Delivery}
To address \textbf{RQ3c}, we examined participants' perceptions and experiences with robot-delivered CBT exercises through post-study interviews.
Despite the LLM system being identical across the chatbot and robot conditions, during our post-study interviews, we observed differences in participants' experiences.

Participant in both conditions reflected positively on the platform convenience, but the students in the robot conditions emphasized that the robot's physical presence was essential to their feeling of connectedness during the exercises. Participants in the chatbot condition expressed feeling that the interaction lacked social connection, with many participants sharing that they felt the chatbot did not care about them, while participants in the robot conditions were very positive about their daily interactions and relationship with the robot. Furthermore, participants in the robot condition were much less critical of LLM-related issues. Embodied agents can access diverse channels of communication, including non-verbal cues that are not available to text-based systems~\citep{deng2019embodiment}. These cues can elicit social, cognitive, and task outcomes~\citep{mutlu2011designing}, and allow a robot to be perceived as a social actor in providing a human-like conversation~\citep{jung2004effects, bainbridge2008effect}. This was reflected in our interviews, where participants talked about perceiving the robot as a friend/human and specifically linked its non-verbal cues, like breathing and head tilts, to making interactions with it feel like human conversations. It is evident that while text-based chatbots are convenient and can effectively guide users through CBT homework exercises, incorporating a physically embodied agent like an appealing robot can improve the user experience.

Overall, this study found strong promise for SAR-based systems as a viable, effective, and well-received alternative to traditional CBT homework exercises.

\subsection{Limitations and Suggestions for Future Work} 

\subsubsection{CBT Exercise Variety}
We implemented two CBT homework exercises and asked participants to choose one of those two to complete each day. We limited the number of exercises to two to simplify the familiarization process with the system. However, we found that the limited variety of exercises, combined with the expectation of daily use, led participants to become fatigued with the two exercises, negatively impacting their adherence to the daily sessions. Future studies should balance the number of activities with the frequency of use and duration of the study, taking care to avoid making participants repeat the same exercises many times. Shorter, flexible sessions can also positively impact adherence, as we observed in our study.

\subsubsection{Treatment Fidelity}
The LLM-enabled exercises were guided by a single system prompt, with no programmed mechanism to enforce the completion of every specific CBT step in every session. Two clinicians' review of a subset of participant transcripts confirmed that while the LLM stayed aligned with the spirit of the CBT exercises, it frequently pivoted to address immediate participant needs, such as providing psychoeducation or rapport-building, rather than strictly following the worksheet's structure. Future work would benefit from implementing state-tracking moderators to ensure CBT exercise alignment in real-time.

\subsubsection{Clinical Context of CBT Homework}
The CBT exercises used in this work did not fully reflect the context in which CBT homework is typically administered. While there exist self-guided CBT homework exercises developed for those without access to clinician support (for example, therapy apps on smartphones, self-help books, etc.), in clinical practice, CBT homework exercises are commonly accompanied by therapist guidance and feedback. By presenting the exercises as a standalone intervention, the study excluded components that may contribute to the effectiveness of CBT-based practices. Consequently, this study should not be interpreted as representing a complete clinical CBT intervention. Rather, the goal of this study was to compare different modalities for delivering CBT homework activities within an exploratory research context.
Future work would benefit from investigating therapist-in-the-loop approaches that better reflect current clinical practices.

\subsubsection{Sample Size}
A primary limitation of the described study is its small sample size ($N=38$), with 12–14 participants per condition. The sample size of this study is consistent with a large body of work in human-robot interaction that is constrained by the challenges of scaling up experiments with real robot hardware (as discussed in Section 2.3).  Consequently, this work should be viewed as an exploratory study rather than a definitive hypothesis-testing experiment. We implemented a Mixed-Effects Model to account for the nested nature of the data and maximize statistical power, but future large-scale longitudinal studies are required to confirm these preliminary trends.

\subsubsection{Study Duration}
Although it has been shown that even short-term CBT interventions can yield positive outcomes~\citep{bertuzzi2021single}, and immediate gains following single-session interventions may predict longer-term symptom improvement~\citep{schleider2019immediate}, the 15-day duration of our study is substantially shorter than typical CBT programs, which often span 6-10 weeks or longer and are commonly delivered through weekly sessions rather than near-daily exercises~\citep{ruwaard2012effectiveness}. While our study participants were exposed to a substantial amount of CBT content, the compressed schedule may have influenced how they engaged with the exercises and limited opportunities to reflect on and integrate newly learned skills into their daily lives. As such, it is difficult to determine whether the observed effects reflect lasting therapeutic benefits, novelty effects associated with the technology, or factors related to participants' day-to-day circumstances. Future work should use longer study periods.

\subsubsection{Interaction Design}
This study was one of the first~\citep{OConnell2024DesignAE} to use the Blossom research robot in a real-world in-home setting, deploying it in university dorms. As anticipated, there were some logistical difficulties associated with the non-commercial research SAR system. In future work, we aim to improve the system design to make it more robust for in-home use. Following participants' suggestions, we hope to explore how to create a more immersive interaction between the robot and the user, for example by enabling the robot to not only suggest coping strategies, but to guide the user through an example strategy, such as meditative breathing. 

Participants wanted the SAR or chatbot to recall information that they had previously disclosed so they could pick up topics from previous sessions and avoid repeating context information in each session. Future systems should incorporate basic strategies for between-session memory recall~\citep{hatalis2023memory}. Future work should also study the effect of between-session memory recall on users' perceptions of mental health agents and therapeutic outcomes.

\subsubsection{Personalization}

Participants in our study expressed a desire to personalize the robot's appearance,  consistent with prior findings showing that robot personalization along several dimensions can improve user perceptions~\citep{shi2023evaluating,sung2009pmroomba,lee2012personalization}. 
We observed a diverse set of preferences and expectations among the participants around the robot's voice and movement. 
We suggest that, when possible, future SAR deployments consider personalizing the robot to fit the preferences of each user, allowing them to control several aspects of the robot, including the color and texture of the exterior, the perceived age and gender of the voice, and the type of movements the robot performs during the interaction. 

\subsubsection{Isolating Embodiment Effects}
In comparing the robot condition to the chatbot condition, two variables were introduced: physical embodiment and auditory output. The robot verbalized responses using a text-to-speech system, while the chatbot was entirely text-based. This design choice was made because disembodied voices in standard chat interfaces can be perceived by users as unnatural, and this creates a confound preventing us from definitively isolating the effect of physical embodiment. Future explorations could include a voice-enabled chatbot or a screen-based virtual agent as an additional control condition.
\subsubsection{Repeated Measurement Effects}
Participants completed the same anxiety instrument before and after each session across all study days. Repeated administration of the measure may have introduced a test–retest effect, whereby participants’ responses could have been influenced by familiarity with the instrument or response habituation over time, potentially contributing to reductions in reported anxiety across the study period.

%% file: sections/statements.tex
\section*{Statements}

\subsection*{Data availability statement}
The datasets presented in this article are not readily available due to the sensitive nature of the collected data, and our IRB approved consent forms. Requests to access the datasets should be directed to MK, kian@usc.edu.

\subsection*{Ethics statement}
The study involving humans were approved by University of Southern California Social Behavioral Institutional Review Board. The study were conducted in accordance with the local legislation and institutional requirements. The participants provided their written informed consent to participate in this study.

\subsection*{Author Contributions}

MK - 
Conceptualization,
Data Curation,
Formal Analysis,
Investigation,
Methodology,
Project Administration,
Resources,
Software,
Supervision,
Writing.

MZ - 
Conceptualization,
Data Curation,
Formal Analysis,
Software,
Visualization,
Writing.

KF - 
Data Curation,
Formal Analysis,
Investigation,
Methodology,
Writing.

AS - 
Data Curation, 
Formal Analysis,
Investigation,
Software,
Visualization,
Writing.

AMV -
Conceptualization,
Formal Analysis,
Investigation,
Methodology,
Writing.

AO - 
Writing – Review \& Editing.

LI -  
Writing – Review \& Editing.

PS - 
Investigation,
Software,
Writing.

SU - 
Investigation,
Software,
Writing.

AG - 
Investigation,
Software,
Writing.

MAF - 
Investigation,
Writing.

WB - 
Investigation,
Software,
Writing.

SMRA - Conceptualization, 
Writing – Review \& Editing.

MC - Formal analysis, Writing – review and editing.

BK -  Conceptualization, Funding Acquisition, 
Writing – Review \& Editing.

MM - Conceptualization, Funding Acquisition, Resources, 
Writing – Review \& Editing.

\subsection*{Funding}
The author(s) declared that financial support was received for this work and/or its publication. Research reported in this publication was supported in part by the University of Southern California (USC) Viterbi School of Engineering, which supported faculty salaries and infrastructure, and the Center for Undergraduate Research in Viterbi Engineering (CURVE) program, which supported undergraduate researcher salaries; the National Science Foundation (NSF) under grant number: NSF IIS-1925083, which supported the development and construction of the Blossom robots deployed in the user study; and the National Institute of Mental Health of the National Institutes of Health (NIH) under Award Number R01MH139134, which supported partial faculty summer salary, partial PhD student salary, and publication costs. Of the approximate \$93,000 spent on the project, an estimated 90\% of the costs were financed by non-governmental sources, an estimated 7.8\% of the costs were financed by NIH, and 2.2\% were financed by NSF.

\subsection*{Conflict of interest}
The author(s) declared that this work was conducted in the absence of any commercial or financial relationships that could be construed as a potential conflict of interest.

%% file: sections/appendix.tex
\section{Supplementary Materials}
\subsection{Race/Ethnic Group Distribution}

\begin{figure}[h]
\centering
\includegraphics[width=0.8\linewidth]{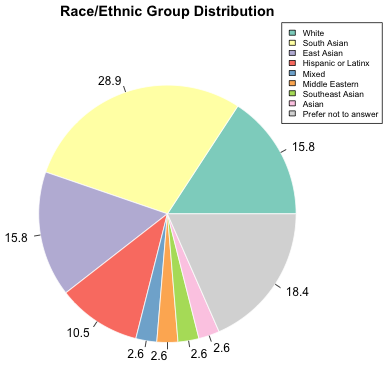}
\caption{Race/Ethnic Group Distribution Figure}
\label{fig:race_distribution}
\end{figure}

\subsection{Blossom Robot Design}

\begin{figure}[h!]
\centering
\includegraphics[height=0.4\columnwidth]{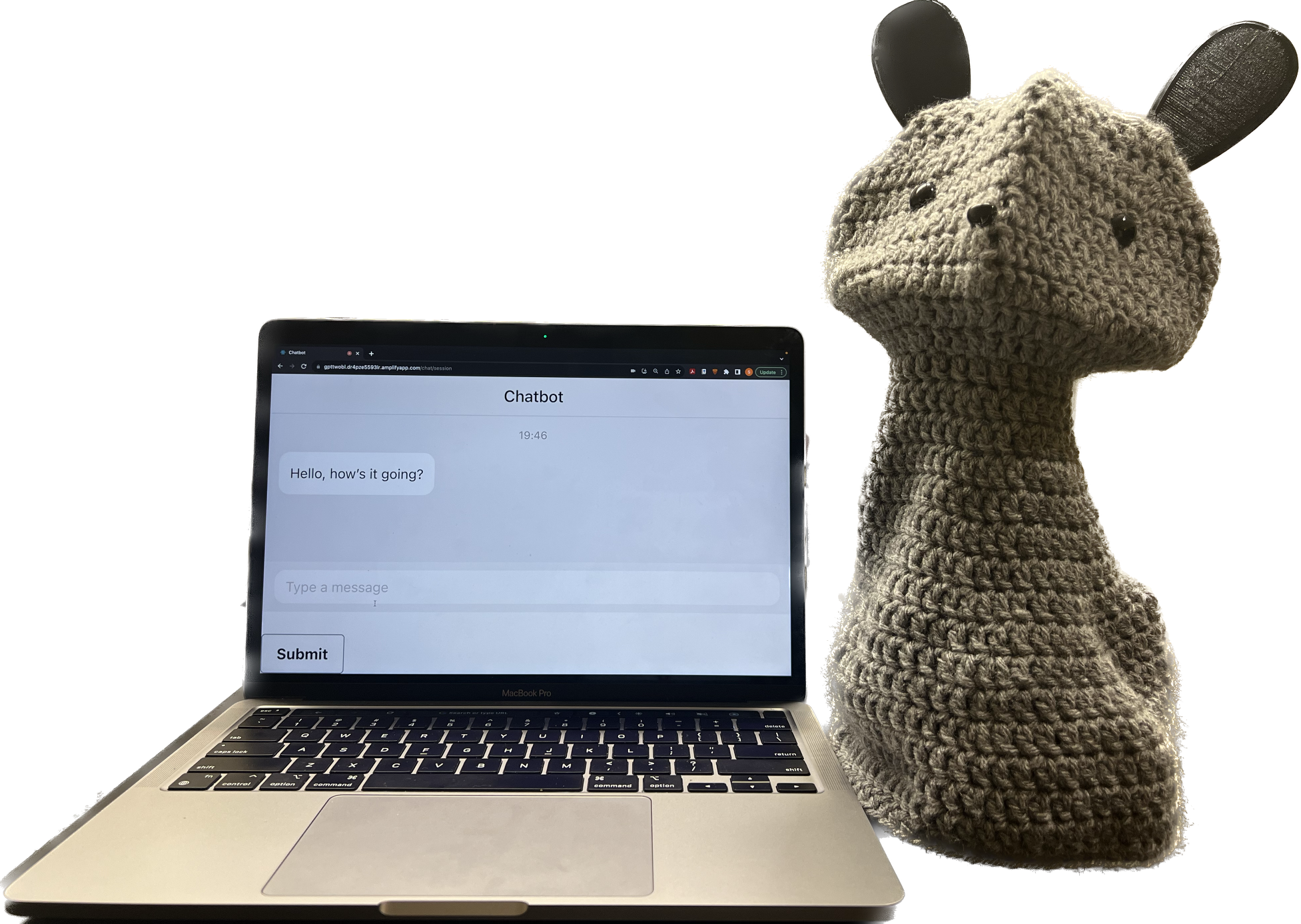}
\caption{Blossom Robot Design}
\label{fig: Blossom_Robot_Design}
\end{figure}

\subsection{GPT-3.5 Parameters}
The input parameters for OpenAI’s chat completion API were:\\
\begin{table}[h]
\begin{tabular}{|l|l|}
\hline
model & gpt-3.5-turbo \\ \hline
messages & $<$transcript$>$ \\ \hline
stop & Patient \\ \hline
temperature & 1 \\ \hline
frequency\_penalty & 2 \\ \hline
presence\_penalty & 2 \\ \hline
n & 2 \\ \hline
max\_tokens & 150 \\ \hline
\end{tabular}
\caption{Parameters for OpenAI's completion API}
\label{tab:model_parameters}
\end{table}
The transcript being sent includes all of the previous messages and the user's latest message.

For moderator, the input parameters were same as above, except the temperature was increased to 2 and the values for n and max\_tokens were not sent.

\subsection{Libraries used for Data Analysis}

Python libraries and their versions:
\begin{table}[h]
\begin{tabular}{|l|l|}
\hline
pandas & 2.0.2 \\ \hline
matplotlib & 3.7.1 \\ \hline
scipy & 1.10.1 \\ \hline
pingouin & 0.5.3 \\ \hline
numpy & 1.24.3 \\ \hline
\end{tabular}
\caption{Python libraries}
\label{tab:python_packages}
\end{table}
R packages:
\begin{table}[h]
\begin{tabular}{|l|l|}
\hline
readxl & 1.4.2 \\ \hline
dplyr & 1.1.2 \\ \hline
tidyr & 1.3.0 \\ \hline
outliers & 0.15 \\ \hline
rstatix & 0.7.2 \\ \hline
ggpubr & 0.6.0 \\ \hline
car & 3.1-2 \\ \hline
lmtest & 0.9-40 \\ \hline
DescTools & 0.99.49 \\ \hline
ggplot2 & 3.4.2 \\ \hline
ggcorrplot & 0.1.4 \\ \hline
MASS & 7.3-60 \\ \hline
afex & 1.3-0 \\ \hline
\end{tabular}
\caption{R packages}
\label{tab:r_packages}
\end{table}

\section{Cognitive Restructuring Worksheet}

\includegraphics[scale=0.6,page=1]{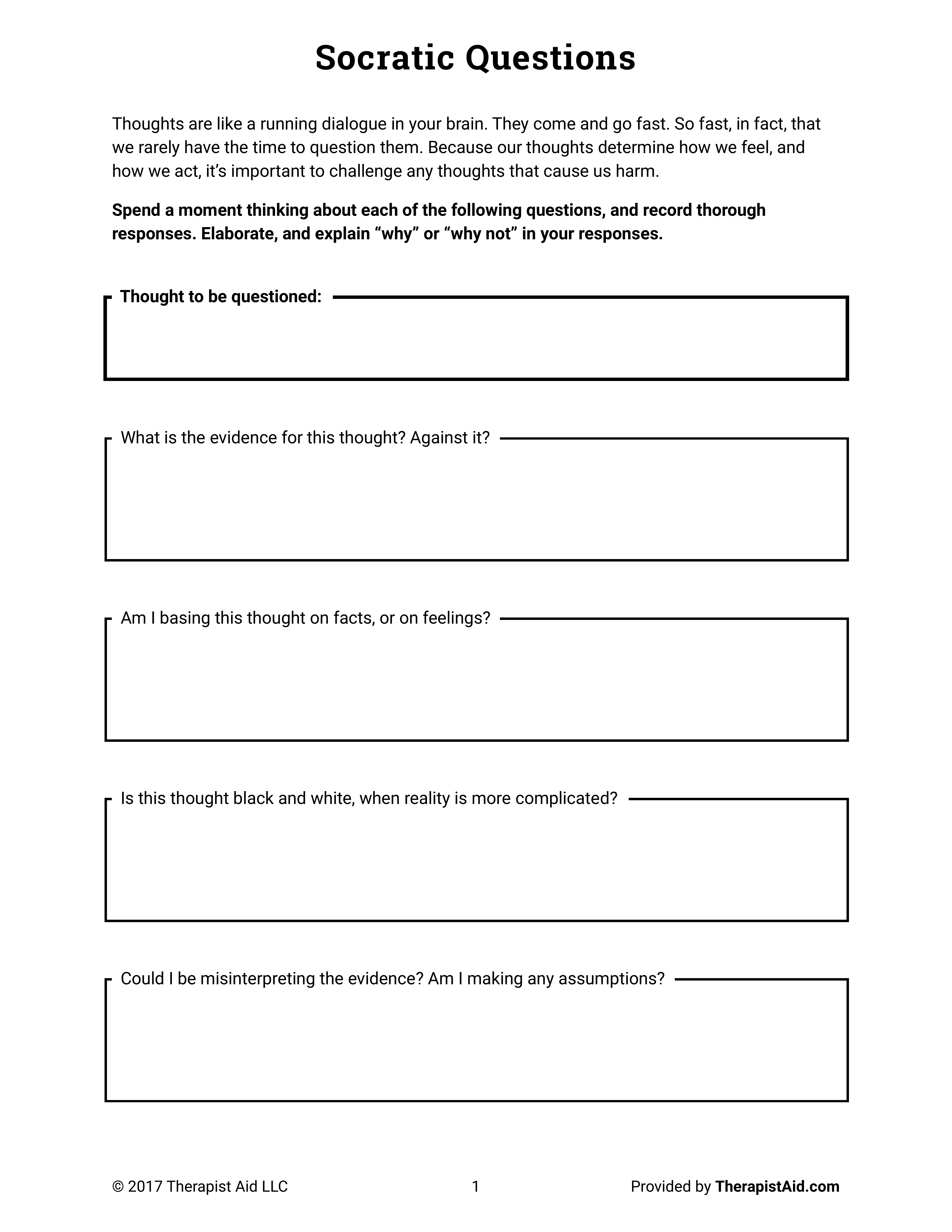}

\section{Coping Strategies Worksheet}

\includegraphics[scale=0.6,page=2]{images/Coping-Styles-Formulation.pdf}

\section{Extended Thematic Analysis}
\subsection{Augmenting Therapy: The Potential Role of SAR CBT Companions in Therapy Practice}

\textbf{Worksheet}

Participants in the worksheet conditions emphasized the importance of anonymity in completing the worksheets. Some described feeling uncomfortable opening up to other people, so they were more comfortable writing about their thoughts by themselves. Similarly, they recalled being able to share their thoughts more freely knowing that no one was listening to them and they were not being judged for the things they wrote: P36 appreciated being able to \textit{`` share without judgement [...] knowing it was confidential but also not having someone there expecting my answers.’’} One participant suggested that items written down in private are more truthful than when said to a person. A few participants noted some drawbacks to filling out a worksheet as an individual, stating that writing their feelings in a worksheet felt one-sided. P36 expressed a desire to be understood that could not be fulfilled by a worksheet \textit{``I'd prefer to talk to a real person though because..[this] technology has never experienced real life."} One participant wanted to know how many other people are dealing with the same problems as them.

\textbf{Chatbot}

One theme that emerged from the chatbot condition is convenience. Participants appreciated that they could use the platform anytime they needed, making the platform good for immediate help. P17 stated, \textit{``[the chatbot] allows you to talk about immediate stress. Couldn't do [that] with a therapist.”}, and P16 observed that \textit{``the chatbot aspect makes it convenient. You can pull it up whenever you need it. Since there wasn't a set time, it was convenient for me.”} They shared that the exercises fit into a busy schedule \textit{``really easy, because it would only take like 10 mins to check in, especially on a day which was bad”} (P22).

Many participants in the chatbot condition shared that the chatbot was good for completing at-home exercises and addressing minor concerns, but emphasized that it was not a replacement for therapy with a human being: \textit{``For people who aren't in therapy for bigger things, lower level issues, it makes sense that this therapy could be accessible and administered through chatbot."} (P22). Two participants stated that they preferred interacting with a human therapist and had no need for a chatbot.

Like participants in the worksheet condition, a few shared that they felt no hesitation to write about their thoughts because they weren’t interacting with a human who could judge them. P26 described the chatbot as \textit{‘``more objective than a human.”}

\textbf{Blossom}

A few participants shared that Blossom could be helpful in conjunction with a human therapist, but was not an alternative to CBT with a person. They appreciated that the robot adjusted its personality and conversation style to match theirs in a way that humans typically don’t, also noting that they felt more comfortable using casual language and punctuation (like texting) with Blossom than they would with a therapist. Additionally, one participant appreciated that the robot could be used without health insurance. 

Participants made several comments about the convenience of the robot-guided CBT homework exercises. They spoke positively about being able to complete the exercises at home any time they wanted, without needing to make an appointment. P14 commented \textit{``Having access to a pseudo-therapist was nice, someone I could talk to whenever I want was definitely a highlight’}. P8 compared it to going to see an in-person therapist, stating, \textit{``[in-person] therapy takes a lot of effort (schedule, pay, go to place). It takes a lot of time, so Blossom offers a place to express emotions immediately.”} They also appreciated the flexible duration of the interactions. However, one challenge that participants faced was being able to find a private time to complete the interaction, as many of them shared a bedroom with a roommate. 

As in the worksheet and chatbot conditions, several participants in the robot condition appreciated that they could talk to Blossom without the fear of being judged that comes with interacting with a human. P7 stated that the robot \textit{``made me feel like I could talk about things without judgement. I think it's nice to know that the view that the AI has will be mostly objective.”} P3 recalled that, at first, they \textit{``didn't want to talk about anything too personal [...] But later [...] I knew it wouldn't judge me [...] I think people are wired to have some biases but this is a robot without emotions that doesn't really think or judge.”} P10 commented on the accessibility this technology could bring to mental health support, suggesting that Blossom could \textit{``encourage [people] to get therapies in ways they otherwise might not, especially if they're introverted or antisocial. It would be good for them."} A few participants recalled bringing their issues to Blossom when their friends didn’t have time to talk without having to worry about wasting Blossom’s time. However, upon reflecting about discussing their issues with Blossom, one participant expressed shame about having to talk to a robot because they didn’t have any close friends to confide in. 

\subsection{General Comments}

\textbf{Increased Guidance}

A few participants across the three conditions wished they had received more guidance on how to complete the exercises. Participants across all three conditions expressed wanting more variety in the exercises provided, instead of having to choose between two. P17 (chatbot) stated \textit{``Only having the 2 exercises was kind of limiting; there's only so much you can do with coping mechanisms and cognitive restructuring.”} P32 (worksheet) suggested \textit{“doing more than 2 options for the exercise [...] the two were great but [...] [the activity would be better] if there was more variety.”} They shared that completing these exercises became repetitive by the end of the study. This feedback was especially prevalent in the worksheet condition, where the participants expressed frustration with filling out the same questions every day.

\textbf{Interaction Design}

Participants in all conditions expressed that the website was easy to navigate and did not have any issues. Some shared feedback about the interaction design.

One participant in the robot condition expressed wishing that they could verbalize their responses instead of typing them out, while another participant appreciated how typing encouraged reflection. Participants in the robot and chatbot conditions wished that they could practice the suggested coping strategies during the interaction. P6 (robot) suggested \textit{``When [the agent] gives a list of potential ways to practice this thought pattern or solution, it might be helpful to offer to practice it."} A few participants in the chatbot condition observed that two weeks may not be enough time to observe changes in mental health or behavior. They had mixed thoughts on the duration of the interaction: some reported that a 30-minute interaction was too long and interrupted their day, while others thought the duration was ok and appreciated having control and flexibility in the duration of their interaction with the chatbot. Participants across all conditions expressed that they appreciated the daily reminders to complete the exercises. 

\textbf{Pre- and Post-Surveys}

Some participants across conditions found that completing the pre- to post-session surveys was helpful for self-reflection. They shared that they were able to recall their previous answers and could use it as a measure of how much their feelings had improved as a result of the interaction. \textit{“It is good to check in on thoughts, pre and post surveys were helpful”} (P14, robot);  \textit{“The survey questions enhanced my ability to dig up feelings”} (P32, worksheet). However, a few participants shared that they found the surveys to be too long and that they disincentivized them from completing the interaction. P24 explained, \textit{``I wanted to talk to [the chatbot] once in the second week, but it would be annoying to fill out the surveys, so it was not worth the effort.”}

\textbf{Participant Comfort over Time and Privacy Concerns}

Across all three conditions participants shared concerns for their privacy, wanting \textit{‘‘better knowledge of what [the researchers] are doing with the surveys and data’’} (P31, worksheet). A few participants in the robot condition shared that because of these concerns they were less vulnerable in their discussions. A theme that emerged across all the conditions was that over time, participants became more comfortable writing their thoughts. Participants in the chatbot condition shared that they opened up more over time. This sentiment was most prevalent in the robot condition. Participants shared that while they were initially hesitant, they became more comfortable and open over time. One participant said they felt comfortable knowing what to expect from the system, and several others shared that becoming familiar with the system allowed for deeper reflection.

Participants in the chatbot and robot conditions shared that they were initially unsure what to expect, with some coming into the study with low expectations: \textit{``I didn't really have a way to navigate what to say. I was definitely holding back [...] It was definitely a fear of the unknown for sure."} (P28, chatbot). However, they shared that both systems exceeded their expectations. Some participants in the chatbot and robot conditions shared that it took time to learn to interact with the chatbot, and several others shared that becoming familiar with the system allowed for deeper reflection. P6 (robot) recalled \textit{``"The interactions became a lot more organic for me […] Over time it got a lot more casual. I got less strict with myself [...] realizing I really can ask anything I want to and it would be fine".} Participants in all three conditions reported becoming more open and comfortable disclosing personal thoughts to their respective platforms over time. P4 (robot) stated \textit{``As I was moving through [the] 7 days, I was opening up a little more each day.”} P28 (chatbot) recalled, \textit{``I struggle with body positivity in a lot of areas. I didn't share that with the chatbot [at first] but then [I] started to share it."} P7 (robot) shared that the topics of discussion became more intimate over time, \textit{``It was easier for me to talk to the robot on day 8 rather than day 1. [At first] I didn't know what to tackle or talk about. I just talked about surface-level stresses like financial aid or family interaction as opposed to later [on, I talked about] about my struggles in isolation and connection."}